\documentclass[11pt]{article}

\usepackage[final]{acl}

\usepackage{times}
\usepackage{latexsym}

\usepackage[T1]{fontenc}

\usepackage[utf8]{inputenc}

\usepackage{microtype}

\usepackage{inconsolata}

\usepackage{graphicx}

\usepackage{amsmath,amsfonts,bm}
\usepackage{amsthm,amssymb}
\usepackage{mathtools}
\usepackage{multirow}

\newtheorem{remark}{Remark}[section]
\newtheorem{proposition}{Proposition}[section]
\usepackage{adjustbox}
\usepackage{subcaption}
\usepackage{cleveref}
\usepackage[linesnumbered,ruled,vlined]{algorithm2e}

\usepackage{soul} 

\def\eqref#1{equation~\ref{#1}}

\def\1{\bm{1}}

\DeclareMathAlphabet{\mathsfit}{\encodingdefault}{\sfdefault}{m}{sl}
\SetMathAlphabet{\mathsfit}{bold}{\encodingdefault}{\sfdefault}{bx}{n}

\newif\ifappendix
\appendixtrue

\title{Principled Detection of Hallucinations in Large Language Models via Multiple Testing}

\author{
 \textbf{Jiawei Li\thanks{Equal contribution.}\textsuperscript{1}},
 \textbf{Akshayaa Magesh$^*$\textsuperscript{2}},
 \textbf{Venugopal V. Veeravalli\textsuperscript{1}}
\\
 \textsuperscript{1}University of Illinois Urbana-Champaign,
 \textsuperscript{2}Meta
\\
  \small{
   {\texttt{\{jiaweil9,vvv\}@illinois.edu}, \texttt{akshayaamagesh@gmail.com}}
 }
}

\begin{document}
\maketitle
\begin{abstract}
While Large Language Models (LLMs) have emerged as powerful foundational models to solve a variety of tasks, they have also been shown to be prone to \emph{hallucinations}, i.e., generating responses that sound confident but are actually incorrect or even nonsensical. 
Existing hallucination detectors propose a wide range of empirical scoring rules, but their performance varies across models and datasets, and it is hard to determine which ones to rely on in practice or to treat as a reliable detector.
In this work, we formulate the problem of detecting hallucinations as a hypothesis testing problem and draw parallels with the problem of out-of-distribution detection in machine learning models. 
We then propose a multiple-testing-inspired method that systematically aggregates multiple evaluation scores via conformal p-values, enabling calibrated detection with controlled false alarm rate.
Extensive experiments across diverse models and datasets validate the robustness of our approach against state-of-the-art methods.
\end{abstract}

\section{Introduction}
Large language models (LLMs) ~\citep{caruccio2024claude,team2024gemini,achiam2023gpt} have emerged as powerful tools for a variety of tasks, most commonly generating answers to user-specified prompts, including text generation, summarization, and question answering~\citep{keskar2019ctrl,raffel2020exploring,zhang2024benchmarking,singhal2025toward}. Although LLMs have shown strong capabilities in these applications, they have also been shown to be prone to \emph{hallucinations}, i.e., generating responses that sound confident, but are actually incorrect, or even nonsensical \citep{ji2023survey, yao2023llm, zhang2023language}. 
Given the increasing reliance on LLMs in real-world scenarios, it is imperative to develop methods to detect whether an LLM is generating hallucinations for a given prompt. 

The term `hallucination' is general, and it captures different kinds of incorrect generations induced by different causes. For example, hallucinations can be classified as factual hallucinations and faithful hallucinations based on the error in the words ~\citep{huang2025survey}. The causes of hallucinations vary, such as LLMs not learning the knowledge to answer the questions ~\citep{onoe2022entity}, LLMs being trained on biased data~\citep{ladhak2023pre}, or LLMs producing sycophancy resulting from reinforcement learning from human feedback (RLHF) training models~\citep{sharma2023towards}. Apart from these, confabulations, i.e., arbitrary and possibly wrong generations from the language model caused by sensitivity to hyperparameters such as random seed, are also an important factor that can lead to incorrect answers in LLMs~\citep{farquhar2024detecting}.

Various approaches have been proposed on hallucination detection in LLMs. External knowledge retrieval approaches ~\citep{chern2023factool,huo2023retrieving} contrast model outputs with external databases to flag factual inconsistencies. Methods leveraging natural language inference (NLI) frameworks ~\citep{zhou2020detecting} assess the consistency between generated content and canonical answers or reference facts, providing another lens to evaluate hallucinations. More recently, the `LLM-as-a-judge' paradigm has emerged, wherein fine-tuned LLMs are tasked with judging the veracity of their outputs. Some studies have explored direct confidence scoring of model generations ~\citep{luo2023chatgpt}, while others have proposed frameworks that design prompts with diagnostic questions to probe hallucinations ~\citep{manakul2023selfcheckgpt,muhammed2025selfcheckagent,yang2025hallucination}. Complementing these approaches, uncertainty estimation methods have been introduced to quantify the inherent ambiguity in model predictions – a factor often observed in hallucinated outputs~\citep{varshney2023stitch,manakul2023selfcheckgpt, rateike2023weakly}. 

It remains unclear
whether any one of these evaluation methods is sufficient to detect all classes of hallucinations. Hallucination patterns can manifest in different ways, making it unlikely that a single metric can detect hallucinations across diverse datasets and models. Additionally, given the rapid pace with which new LLMs are being developed, it is not realistic to develop specific hallucination detection methods for each new LLM.  Therefore, it is of interest to design a generally robust method for hallucination detection that can leverage the advantages of preexisting methods, without any additional assumptions on the specific datasets or the LLMs being considered.

In this work, we build a unified multiple testing framework for hallucination detection that systematically combines scores proposed in prior work. The framework is agnostic to how the underlying scores are obtained; in principle, it can incorporate any collection of baseline detectors. In our experiments, however, we focus on training-free, retrieval-free scores that require no external factual repositories and no auxiliary judge models. Concretely, we leverage state-of-the-art uncertainty and consistency metrics derived from the model’s output layer by resampling multiple generations under different random seeds. For example, \citet{kuhn2023semantic} propose semantic entropy, which leverages meaning-invariant clustering to quantify uncertainty, and \citet{lin2023generating} propose confidence measures computed from pairwise similarity among sampled generations, including lexical overlap and spectral scores based on a similarity graph. Finally, our framework adds a lightweight calibration step that uses a small calibration set of non-hallucinated prompts to provide theoretical control of the false-alarm rate.

\subsection{Our contributions}
We develop a robust method for hallucination detection by framing it as a hypothesis testing problem. 
Motivated by recent advances in out-of-distribution (OOD) detection, we propose adapting the principled detection procedure developed by \cite{magesh2023principled} to the problem of hallucination detection in LLMs.
Specifically, we introduce a method that systematically integrates multiple evaluation scores using conformal p-values. 
Our key contributions are summarized as follows.

\paragraph{1. A hypothesis-testing-based framework.} We reconceptualize hallucination detection as a hypothesis testing problem, drawing parallels with OOD detection in machine learning (Section~\ref{sec: problem modeling}). This provides a statistically grounded framework for identifying hallucinations in LLM-generated content.

\paragraph{2. A multiple-testing-inspired detection pipeline.} We provide a non-trivial conceptual bridge that adapts multiple hypothesis testing to the problem of LLM hallucination. Unlike existing ad-hoc heuristics, our framework systematically manages the dependencies among heterogeneous signals, ranging from lexical overlap to deep semantic and spectral properties. This represents a significant shift toward principled inference, providing the first framework capable of offering calibrated false-alarm control in a generative context (Section~\ref{sec: method}).

\paragraph{3. Empirical validation across diverse datasets and models.}  We conduct extensive experiments across various LLM architectures and datasets (Section~\ref{sec:experiment}), demonstrating that our method maintains robustness across different datasets and LLMs, and generally outperforms existing hallucination detection techniques. Specifically, our method exhibits consistently high AUROC and detection power across LLaMA-2, LLaMA-3, Mistral, and DeepSeek-v2, on HaluEval, CoQA and TriviaQA (\Cref{tab: auroc of coqa and trivia qa(10 repeated exps)}), whereas state-of-the-art methods exhibit high variability among different models and datasets. 
Moreover, when we macro-average AUROC over the full evaluation suite (spanning all tested models and datasets, including additional models such as Qwen3 variants and additional benchmarks such as FactCHD and GSM8K), our method attains the highest average performance, outperforming both prior baselines and simple aggregation strategies, such as majority voting and averaging (Figure~\ref{fig: average AUROC}).
These empirical results corroborate the effectiveness and robustness of our proposed approach.

\section{Problem Modeling}
\label{sec: problem modeling}
In this section, we first review existing methods in the literature for detecting hallucinations. We then define the problem from the perspective of hypothesis testing and introduce the multiple testing framework to leverage baseline methods. The objective is to distinguish prompts that are less likely to generate hallucinations, labeled as \emph{correct}, from prompts likely to generate hallucinations, labeled as \emph{incorrect}.

\subsection{Existing hallucination detection methods}
\label{sec: baseline scores}
In this part, we introduce several threshold-based methods to detect hallucinations in gray-box settings (with access to output likelihoods) or black-box settings (with access to the generations only).  All these methods develop metrics to measure the uncertainty or similarity of multiple generations given the same prompt, and declare a prompt likely to generate hallucinations based on an empirical threshold applied to the measured metric.

\textbf{Semantic Entropy.} \cite{farquhar2024detecting} consider the entropy of model generations under semantic clustering, where generations with the same semantic meaning are grouped into the same cluster. The  semantic entropy score is calculated as \[\textbf{SE}(x) = - \sum_{i=1}^{|C|} \mathbb P(C_i\;|\;x)\log \mathbb P(C_i\;|\;x), \] 
where $x$ denotes the prompt, $C_i$ denotes the $i$-th semantic cluster, obtained via semantic equivalence clustering based on bidirectional entailment, and $\mathbb P(C_i\;|\;x)$ is the normalized probability mass assigned to cluster $C_i$ conditioned on $x$.

\textbf{Alpha Semantic Entropy (\textbf{$\alpha$SE}).} \cite{kaur2024addressing} also focus on semantic similarity among different generations, but provide a different algorithm to calculate the clustering of semantic equivalence, inspired by the distance-dependent Chinese restaurant process.

\textbf{Spectral Eigenvalue.} Instead of considering the semantic similarity problem as a black-and-white problem, \cite{lin2023generating} consider the semantic similarity between different samples as continuous real numbers and translate them into weights in a graph. The eigenvalues of the symmetric normalized graph Laplacian are then calculated from the symmetric weighted adjacency matrix. Suppose the eigenvalues are $\lambda_1<\cdots<\lambda_m$, then the spectral eigenvalue score is given by
\[\textbf{EigV} (x) = \sum_{i=1}^m \max(0,1-\lambda_i).\]

\textbf{Kernel Semantic Entropy.} \cite{nikitin2024kernel} kernelize semantic entropy using the pairwise semantic similarities among $M$ sampled generations. Let $s_{ij}$ be the symmetric NLI entailment score between generations $i$ and $j$, define $K_{ij}=\exp\!\big(s_{ij}/\tau\big)$ with $\tau>0$, and use weights $w_j\ge 0$ with $\sum_{j=1}^M w_j=1$. Let $p_i=\sum_{j=1}^M w_j K_{ij}$. The kernel semantic entropy is
\[
\textbf{KSE}(x)=-\sum_{i=1}^M w_i \log p_i .
\]
Larger KSE indicates greater semantic dispersion (less clustering) among generations.



\textbf{Lexical Similarity (LS).} The lexical similarity score uses the sum of Rouge-L similarity scores among different samples in the generation, which is proposed in~\cite{fomicheva2020unsupervised}.
\subsection{Multiple hypothesis testing framework}

Consider the question answering scenario where, given an input (prompt) $X$, the goal is to predict the output (answer) $Y$, which follows a ground truth distribution given $X$, $Y \sim P(Y|X)$. \footnote{Throughout the paper, uppercase characters (e.g., 
$X$ and $Y$) denote random variables, while their realizations are denoted by lowercase characters (e.g., $x$ and $y$).} 
In practice, Large Language Models (LLMs) are utilized to approximate this distribution by generating text conditioned on the input $X$. Let the model be denoted as $f(\mathbf W, \cdot)$, where $\mathbf W$ represents the parameters of the LLM. 
Given an LLM and test data $\mathcal T=\{(x_i,y_{i})\}_{i=1}^n$, where $x_i$ is the prompt and $y_{i}$ is the reference answer, the objective is to detect whether the LLM may generate incorrect responses (hallucinations) for the given prompt $x$.
The hallucination detection problem can be posed as a hypothesis testing problem as follows,
{\begin{align*}
H_0:\ &X\text{  is likely not to generate a hallucination}\\&\text{at the output of model }f(\mathbf{W},X),\\
H_1:\ &X \text{ is likely to generate a hallucination} \\&\text{at the output of model } f(\mathbf{W},X), 
\end{align*}}
where $H_0$ is the null hypothesis, and $H_1$ is the alternative hypothesis. 

As in general hypothesis testing settings, there are two metrics for performance evaluation, false alarm rate $\mathrm{P_F}$ and detection power $\mathrm{P_D}$. The false alarm rate indicates the probability of misclassifying correct generations from a given prompt as a hallucination, while the detection power is the probability of correctly detecting hallucinations. \footnote{Here $\mathbb P_{H_i} (\text{declare hallucination})$ means that, given the model $\mathbf{W}$ and prompt $X$, under the hypothesis $H_i$, the probability that the method declares that the prompt $X$ will generate hallucinations.}
\begin{align*}
\mathrm{P_F} = &\mathbb P_{H_0} (\text{declare hallucination}),\\
\mathrm{P_D} = &\mathbb P_{H_1} (\text{declare hallucination}).
\end{align*}
As described in the previous subsection, there are works that develop multiple scores for detecting hallucinations. 
Drawing ideas from OOD detection, we propose a multiple testing framework for hallucination detection, where we combine these scores from prior work. Specifically, given the model $\mathbf W$ and the input $X$, a list of scores $T^i$ can be obtained by the score functions,
\begin{align*}
T^1 = s^1(X),T^2 = s^2(X),\cdots, T^K = s^K(X),
\end{align*}
where $s^i(\cdot)$ depends on  the model parameters $\mathbf W$. 
Since $X$ is a random variable, $T^i$ is also a random variable, and we denote the realization of $T^i$ as $t^i$. The multiple testing framework is as follows:
\begin{align*}
H_{0,1}:\ & T^1\sim P^1,&\quad
H_{1,1}:\ & T^1\not\sim P^1;\\
\vdots&&\vdots\\
H_{0,K}:\ & T^K\sim P^K,&\quad
H_{1,K}:\ & T^K\not\sim P^K,
\end{align*}

where $P^1,P^2,\cdots,P^K$ are the induced distributions of the scores for corresponding prompts for which the LLM does not generate hallucinations. We declare a generation from a given prompt to be a hallucination if any of $H_{0,i}$ is rejected, that is, if the \emph{global null} is rejected.

Furthermore, since the distributions $P^1,P^2,\cdots,P^K$ are unknown, an additional calibration dataset $\mathcal{C} = \{x_i^{c}\}_{i=1}^{n_\text{cal}}$ is utilized to provide information about the prompts that lead to correct generations.


\section{Proposed Methodology}
\label{sec: method}
In this section, we present our proposed algorithm for detecting hallucinations in the generations of LLMs. First, we compare the problem of detecting hallucinations in LLMs with the problem of out-of-distribution (OOD) detection in machine learning models, highlighting parallels and differences in Subsection~\ref{subsec: OOD vs HD}. Then, in Subsection~\ref{subsec:proposed_algorithm}, we present our method for detecting hallucinations in LLM generations, adapted from the OOD detection procedure in \cite{magesh2023principled}.

\subsection{Out-of-distribution detection and hallucination detection}
\label{subsec: OOD vs HD}
A multiple hypothesis testing approach  for out-of-distribution detection was first proposed in \cite{magesh2023principled}. Here, we adapt this approach for hallucination detection. The two problems share several key similarities. 
They both aim to detect untrustworthy predictions or generations to improve robustness for the safe deployment of these models.
Neither problem setting imposes assumptions on the distributions under the null and alternative hypotheses, because the input distribution is unknown when utilizing machine learning models during the test period. Additionally, both operate in a zero-resource setting, i.e., no additional data from the test distribution or extra training is used for detection.

Despite these similarities, there are key differences between hallucination detection and OOD detection as follows.
\paragraph{Indicator scores.} 
Unlike standard OOD detection, which typically operates in a white-box setting by utilizing internal activations from multiple intermediate layers to capture fine-grained distribution shifts \cite{lee2018simple, de2025mysteries}, our framework is designed for grey-box and black-box regimes. While OOD methods often derive scores from distance metrics within a unified latent space, LLM hallucination detection must bridge heterogeneous signals, ranging from lexical overlap to deep semantic entropy and spectral properties. Consequently, our scores are derived solely from final-layer output likelihoods or sampled generations, which are influenced by hidden parameters but do not directly expose them. This methodological choice is driven by the extreme scale of LLMs, where the vast number of parameters makes computing scores based on hidden-layer outputs costly or even infeasible. Such a limitation is especially critical for closed-source models, where the proprietary nature does not expose internal parameters, making our reliance on output-level signals essential for real-world deployment.

\paragraph{Data drawn from null hypothesis and alternative hypothesis.}
In OOD detection, the model is typically trained on one dataset and tested on another dataset, since small models can be efficiently trained and evaluated on different datasets, enabling clear hypothesis separation. The data from the training dataset serve as data from the null hypothesis, and the data from another dataset serve as data from the alternative hypothesis.
However, LLMs are pretrained on massive corpora at a high computational cost, and hallucination detection is typically performed on a fixed pretrained model without further training. 
Moreover, since LLMs generate outputs probabilistically from logits, different outputs can be produced for the same input, making it difficult to explicitly label data as coming from the null or alternative hypothesis. In this work, we adopt the Rouge-L score between generations and reference answers to label the generations as non-hallucinated or hallucinated. To accommodate potential rephrasing, we regard the input as not generating hallucinations if only a small fraction of sampled generations are classified as hallucinations among all the generations.
\subsection{Proposed hallucination detection method}{\label{subsec:proposed_algorithm}}

The multiple hypothesis testing algorithm is based on the general version of the Benjamini-Hochberg (BH) procedure that allows the scores to be dependent \cite{benjamini2001}. If the distribution of scores under the null hypothesis is known, the p-values of the observed score $t^j_\mathrm{test}$ can be computed as
\[
q^j\coloneqq\mathbb P_{H_0}(T^j\ge t^j_\mathrm{test}), 
\] 
with corresponding random version for a random test score $T^j_\mathrm{test}$ being denoted by $Q^j$.



However, in our hallucination detection problem, we do not have the distribution of scores under $H_0$, but we have access to the calibration set $\mathcal C$, which contains prompts that do not generate hallucinations. The dataset $\mathcal C$ can be used to compute empirical versions of the p-values, known as \emph{conformal} p-values, of the scores.
Denote the scores in the calibration set as $\{s_{i}^j=s^j(x_i^c)\;:\;x_i^c\in \mathcal C, j=1,2,\cdots,K\}$. Given the test scores $t_\mathrm{test}^j$ and the corresponding random variables $T_\mathrm{test}^j$, the conformal p-values and their random versions, conditioned on the calibration dataset $\mathcal C$, are defined as
\begin{align*} q^j_\mathrm{con} &\coloneqq  \frac{1+|\{i:s_i^j\ge t^j_\mathrm{test} \}|}{1+|\mathcal C|},\\
Q^j_\mathrm{con} &\coloneqq \frac{1+|\{i: s_i^j\ge T^j_\mathrm{test} \}|}{1+|\mathcal C|}.
\end{align*}

Algorithm~\ref{alg: multiple hypothesis testing} describes the method inspired by the BH procedure. The hyperparameter $\epsilon$ is related to the concentration property of the CDF of the Beta distribution, as detailed in Remark~\ref{remark: size of calibration dataset}. A higher score indicates a higher likelihood of being classified as a hallucination.
The conformal p-values for the test data are computed using the scores from the calibration dataset $\mathcal C$ and the scores from the test generations. These conformal p-values are then compared against ranked thresholds for hallucination detection. 
Theoretical guarantees for the false alarm rate can be obtained through the following theorem.


\begin{algorithm}[t]
\small
\SetAlgoLined

\KwIn{Test prompt with generations $x_{\mathrm{test}}$; desired false alarm rate $\alpha$; hyperparameter $\epsilon$; calibration dataset $\mathcal C$ and its scores $\{s_i^j = s^j(x_i^c) : x_i^c \in \mathcal C,\ j=1,\ldots,K\}$.}
\KwOut{Decision on whether the prompt with generations is hallucinated.}

\For{\( j \gets 1 \) \KwTo \( K \)}{
  \(t^j_\mathrm{test} \gets s^j(x_{\mathrm{test}})\)\;
  \(q^j_\mathrm{con} \gets \frac{1+\left|\left\{i:\ t^j_\mathrm{test}\le s_i^j \right\}\right|}{1+|\mathcal C|}\)\;
}

\(\widehat{q}_{\mathrm{con}}^1,\ldots,\widehat{q}_{\mathrm{con}}^K \gets\) sorted values of \(\{q_{\mathrm{con}}^j\}_{j=1}^K\) in ascending order\;

\eIf{
\(\exists\, j \in \{1,\ldots,K\} \text{ such that }
\widehat{q}_{\mathrm{con}}^{\,j}\le
\frac{\alpha}{(1+\epsilon)\sum_{i=1}^K\frac1i}\cdot \frac{j}{K}\)
}{
  \Return{Hallucination}\;
}{
  \Return{No Hallucination}\;
}

\caption{Multiple Hypothesis Testing for Hallucination Detection}
\label{alg: multiple hypothesis testing}
\end{algorithm}


\begin{proposition}[Theorem 2 in \cite{magesh2023principled}]
\label{thm: false alarm rate boundness}
Let $\alpha,\delta\in(0,1)$.
Denote the calibration set as $\mathcal C$. When the size of the calibration set $|\mathcal C|$ is sufficiently large, for a new input $X$ and a learning model $f(\mathbf W, \cdot)$, with probability $1-\delta$, the false alarm rate of Algorithm~\ref{alg: multiple hypothesis testing} is bounded by $\alpha$, i.e.,
\begin{align*}
\mathrm{P_F}(\mathcal C) = \mathbb P_{H_0}(\mathrm{declare\ hallucination}\mid \mathcal C) \le \alpha.
\end{align*}
\end{proposition}
\begin{remark} 
\label{remark: size of calibration dataset}
Let $\epsilon > 0$, and denote by $K$ the number of scores to be integrated and by $\alpha$ the desired false alarm rate. As stated in Lemma 1 in \cite{magesh2023principled}, the size of calibration dataset $|\mathcal C|$ is sufficiently large if for the given $\delta >0$, \begin{align*}
\min_{j=1,2,\cdots,K} I_{(1+\epsilon)\mu_j}(a_j,b_j)\ge 1-\frac{\delta}{K^2},
\end{align*}
where $a_j=\lfloor 
\frac{(|\mathcal C|+1)\cdot\alpha}{(1+\epsilon)\sum_{i=1}^K\frac1i}\cdot  \frac{ j }{K}\rfloor$, $b_j =|\mathcal C|+1-a_j $, $\mu_j = \frac{a_j}{a_j+b_j}$, and $I_x(a,b)$ denotes the CDF of Beta distribution $\mathrm{Beta}(a,b)$ evaluated at $x$.

Importantly, the required calibration size depends only on the parameters $(\alpha, \epsilon, \delta, K)$ and is independent of the specific learning problem being considered or the pair of dataset and model being tested. When the calibration set is small, 
$\epsilon$ can be increased to accommodate the reduced sample size, at the cost of a more conservative decision rule that may reduce the detection power.
\end{remark}

\begin{remark}
 Algorithm \ref{alg: multiple hypothesis testing} requires neither additional model training nor extra data beyond what the underlying baseline scores already use. A calibration set can be constructed from prompts in the training data that are deemed non-hallucinated. As a result, our approach introduces minimal overhead and does not impose additional data or computational requirements beyond those of the chosen baselines. Furthermore, Algorithm \ref{alg: multiple hypothesis testing} is broadly applicable: given a collection of baseline scoring rules for a detection (or more general decision) task, it can aggregate them into a single calibrated test with controlled false alarm rate. 
 \end{remark}

\begin{remark}
Unlike prior work  providing marginal guarantees over random calibration sets, this new method focuses on the conditional false alarm rate, which can be controlled at all times as long as the size of the calibration dataset is sufficiently large. The calibration set’s diversity and coverage mainly affect detection power (e.g., redundancy can reduce power), while the false alarm guarantee remains valid.
\end{remark}

The final part is to construct the calibration dataset, which requires a method to label whether a given prompt with generations is hallucinated or not. 
Using the reference answer, each prompt-generation pair is labeled based on the Rouge-L similarity, following the criteria in \citet{lin2023generating}.  As shown in Algorithm~\ref{alg: classify dataset}, once the Rouge-L similarity scores are high for most generations, the prompt is deemed not likely to generate hallucinations, since its generations are approximately correct. 
Finally, the calibration dataset is constructed by sampling from prompts that do not generate hallucinations. 

\begin{algorithm}[t]
\small
\SetAlgoLined
\KwIn{Dataset $\mathcal D$ with generations $\mathcal D_{p,j}$ for prompt $p$, reference answers $Y$ with $Y_p$, similarity threshold $\tau$, tolerance $\theta \in [0,1]$.}
\KwOut{Prompt sets $\mathcal D_{\mathrm{NH}}$ (non-hallucinated) and $\mathcal D_{\mathrm{H}}$ (hallucinated).}

$\mathcal D_{\mathrm{NH}}, \mathcal D_{\mathrm{H}} \gets \emptyset, \emptyset$\;

\For{$p \gets 1$ \KwTo $n_{\mathrm{prompts}}$}{
  $h \gets \left|\left\{j \in [n_{\mathrm{gen}}] :
  \mathrm{Rouge\text{-}L}(\mathcal D_{p,j}, Y_p) \le \tau\right\}\right|$\;

  \eIf{$\frac{h}{n_{\mathrm{gen}}} \le \theta$}{
    $\mathcal D_{\mathrm{NH}} \gets \mathcal D_{\mathrm{NH}} \cup \{p\}$\;
  }{
    $\mathcal D_{\mathrm{H}} \gets \mathcal D_{\mathrm{H}} \cup \{p\}$\;
  }
}
\Return{$(\mathcal D_{\mathrm{NH}}, \mathcal D_{\mathrm{H}})$}\;
\caption{Assign hallucination labels to prompts}
\label{alg: classify dataset}
\end{algorithm}


\section{Experimental Results}
\label{sec:experiment}
We demonstrate that our method shows robustness consistently across different datasets and language models. That is, regardless of the tested LLM or dataset, our method achieves consistently strong performance on hallucination detection. In contrast, other baseline scores often degrade in performance when faced with particular model–dataset combinations, or fail to outperform our method consistently. This highlights the reliability and generalizability of our approach, especially in real-world scenarios where user queries come from unknown and highly variable distributions. 

\textbf{Experimental setup.} Based on the $20$ sampled generations given the input, the Rouge-L score threshold, $\tau$, is set to $0.3$ to determine whether a generation is considered a hallucination. For a prompt, if at least 18 out of the 20 sampled generations ($\theta= 0.1$) are judged as non-hallucinations under Rouge-L evaluation, the prompt is considered not to generate hallucinations. 
We repeat the experiments $10$ times using a randomized calibration dataset (a subset of prompts with non-hallucinated generations), and report the mean and standard deviation of our evaluation metrics. By default, the size of the calibration dataset is $1,000$.
We test on models such as LLaMA-2-13B~\citep{touvron2023llama},  Mistral-7B~\citep{jiang2023mistral7b}, Llama-3.1-8B~\citep{grattafiori2024llama}, DeepSeek-v2-Lite~\citep{deepseekv2}, 
Qwen3-4B~\citep{yang2025qwen3}, Llama-3.2-3B-Instruct~\citep{grattafiori2024llama}, and Qwen2.5-Math-1.5B-Instruct~\citep{yang2024qwen2}.
Following \cite{kuhn2023semantic}, we evaluate our procedure on the HaluEval~\cite{li2023halueval}, TriviaQA~\citep{2017arXivtriviaqa}, CoQA~\citep{reddy2019coqa}, 
and GSM8K~\citep{cobbe2021gsm8k}. 

\textbf{Baselines.} We adopt five scoring functions to quantify the degree of hallucination induced by a prompt (Section~\ref{sec: baseline scores}). In addition, since the clustered variant of semantic entropy based on the frequencies of cluster assignments~\cite{farquhar2024detecting} can sometimes outperform the original formulation, we also report results for this clustered version.


\textbf{Performance evaluation.}
Following prior work, we report Area Under the Receiver Operating
Characteristic curve (AUROC) in the main text; AUROC summarizes the tradeoff between detection power and false-alarm rate over all decision thresholds. In the appendix, we additionally report
 the detection power at a fixed $10\%$ false alarm rate.
Mathematically, given the calibration dataset $\mathcal C$, the detection power is defined as 
\begin{align*}
\mathrm{P_D}(\mathcal C)
&= P_{H_1}\!\bigl(\text{declare hallucination}\mid \mathcal C\bigr),\\
\text{s.t.}\ \ 
\mathrm{P_F}(\mathcal C)
&= P_{H_0}\!\bigl(\text{declare hallucination}\mid \mathcal C\bigr)= \alpha.
\end{align*}

\begin{table*}[t]
\caption{AUROC (\%) and detection power ($P_D$, \%) at 10\% false alarm rate across different models and datasets. $*$ and $\dagger$ indicate the best and worst performance in each setting, respectively. Boldface highlights our method, which achieves the highest performance in the majority of evaluated cases.}
\centering
\begin{adjustbox}{width=0.9\linewidth}
\begin{tabular}{llcccccccc} 
\toprule
\multirow{2}{*}{Dataset} & \multirow{2}{*}{Method} & \multicolumn{2}{c}{Llama-2-13B} & \multicolumn{2}{c}{Mistral-7B} & \multicolumn{2}{c}{Llama-3.1-8B} & \multicolumn{2}{c}{DeepSeek-v2-Lite} \\
\cmidrule(lr){3-4} \cmidrule(lr){5-6} \cmidrule(lr){7-8} \cmidrule(lr){9-10}
& & AUROC & $P_D$ & AUROC & $P_D$ & AUROC & $P_D$ & AUROC & $P_D$ \\
\midrule
\multirow{8}{*}{ HaluEval } 
& SE & $64.34^\dagger \pm 0.23$ & $41.50^\dagger \pm 0.58$& $60.69^\dagger \pm 0.23$ &$35.68^\dagger \pm 0.57$  &$62.24^\dagger \pm 0.23$ & $36.88^\dagger \pm 0.58$ &$70.36^\dagger \pm 0.23$&  $50.99 \pm 0.38$ \\
& $\alpha$SE & $83.43 \pm 0.29$ & $66.46 \pm 0.87$& $80.36 \pm 0.20$ & $60.95 \pm 0.66$ & $81.80 \pm 0.23$   & $62.97 \pm 0.67$& $84.54 \pm 0.21$  & $68.84 \pm 0.86$ \\
& KSE & $89.41 \pm 0.38$ & $71.21 \pm 1.70$ & $88.85^* \pm 0.26$  & $69.90 \pm 1.07$& $89.32 \pm 0.19$ & $69.35 \pm 0.75$& $90.43^* \pm 0.18$  & $72.27 \pm 1.06$\\
& clustered\_SE & $78.32 \pm 0.34$  & $51.03 \pm 0.84$& $78.04 \pm 0.28$ & $50.06 \pm 0.45$ & $78.82 \pm 0.16$ & $51.99 \pm 0.35$ & $83.49 \pm 0.22$ & $60.70 \pm 0.67$ \\
& $\alpha$\_clustered\_SE & $85.83 \pm 0.31$ & $71.77 \pm 0.76$ & $85.53 \pm 0.16$  & $67.32 \pm 0.51$ & $85.40 \pm 0.13$ & $68.18 \pm 0.46$& $87.48 \pm 0.15$ & $71.32 \pm 0.45$ \\
& EigV & $86.59 \pm 0.39$ & $71.02 \pm 1.00$ & $85.83 \pm 0.20$ & $64.44 \pm 0.85$ & $85.90 \pm 0.22$  & $65.39 \pm 0.66$& $87.16 \pm 0.18$ & $67.49 \pm 0.62$ \\
& LS & $88.77 \pm 0.48$  & $64.81 \pm 3.87$& $85.62 \pm 0.34$  & $45.74 \pm 2.35$& $89.22 \pm 0.33$  & $64.34 \pm 1.36$& $85.19 \pm 0.40$  & $41.28^\dagger \pm 5.64$\\
& \textbf{Ours} & $\mathbf{90.18}^* \pm \mathbf{0.34}$ & $\mathbf{75.70}^* \pm \mathbf{1.19}$ & $\mathbf{88.71} \pm \mathbf{0.24}$ & $\mathbf{71.74}^* \pm \mathbf{0.74}$  & $\mathbf{90.56}^* \pm \mathbf{0.20}$ & $\mathbf{74.93}^* \pm \mathbf{0.63}$ & $\mathbf{90.27} \pm \mathbf{0.16}$ & $\mathbf{76.79}^* \pm \mathbf{0.96}$ \\

\midrule
\multirow{8}{*}{ CoQA } 
& SE & $66.68^\dagger \pm 0.14$ & $38.93^\dagger \pm 0.30$ & $65.41^\dagger \pm 0.26$ & $36.62^\dagger \pm 0.63$& $65.32^\dagger \pm 0.18$ & $37.11^\dagger \pm 0.48$ & $69.05^\dagger \pm 0.25$ & $41.98^\dagger \pm 0.58$  \\
& $\alpha$SE & $85.19 \pm 0.20$  & $65.10 \pm 0.82$& $85.15 \pm 0.23$ & $65.42 \pm 1.06$  & $84.17 \pm 0.18$& $61.96 \pm 0.32$ & $86.55 \pm 0.25$  & $66.73 \pm 0.62$\\
& KSE & $88.72 \pm 0.17$& $70.50 \pm 0.57$ & $88.43 \pm 0.28$ & $68.59 \pm 0.74$& $87.77 \pm 0.18$ & $66.96 \pm 0.80$  & $89.05 \pm 0.26$ & $69.53 \pm 1.01$ \\
& clustered\_SE & $85.56 \pm 0.25$ & $57.05 \pm 0.91$& $85.55 \pm 0.31$  & $58.23 \pm 1.07$ & $85.18 \pm 0.21$ & $58.20 \pm 0.46$ & $86.42 \pm 0.29$& $61.03 \pm 0.69$ \\
& $\alpha$\_clustered\_SE & $89.79 \pm 0.15$& $72.48 \pm 0.71$ & $90.04 \pm 0.21$  & $73.62 \pm 1.04$ & $89.46 \pm 0.13$ & $71.89 \pm 0.39$& $90.80 \pm 0.20$  & $74.78 \pm 0.54$ \\
& EigV & $90.03 \pm 0.15$ & $72.08 \pm 0.62$& $90.28 \pm 0.22$ & $73.49 \pm 0.72$ & $89.79 \pm 0.15$  & $72.07 \pm 0.70$ & $91.06 \pm 0.26$& $74.34 \pm 0.94$  \\
& LS & $88.36 \pm 0.49$& $64.33 \pm 2.66$  & $89.28 \pm 0.25$& $68.25 \pm 0.72$ & $87.87 \pm 0.37$  & $61.92 \pm 1.58$ & $89.83 \pm 0.52$& $70.94 \pm 1.02$  \\
& \textbf{Ours} & $\mathbf{90.85}^* \pm \mathbf{0.11}$  & $\mathbf{74.90}^* \pm \mathbf{0.84}$ & $\mathbf{91.23}^* \pm \mathbf{0.25}$& $\mathbf{75.81}^* \pm \mathbf{1.34}$  & $\mathbf{90.44}^* \pm \mathbf{0.13}$ & $\mathbf{74.23}^* \pm \mathbf{0.36}$& $\mathbf{91.74}^* \pm \mathbf{0.19}$  & $\mathbf{76.45}^* \pm \mathbf{1.47}$ \\

\midrule
\multirow{8}{*}{ TriviaQA } 
& SE & $82.52^\dagger \pm 0.13$ & $67.96 \pm 0.35$ & $82.12^\dagger \pm 0.13$ & $67.01 \pm 0.44$& $84.67^\dagger \pm 0.17$  & $69.02 \pm 0.36$ & $87.34^\dagger \pm 0.18$ & $75.43 \pm 0.34$\\
& $\alpha$SE & $90.75 \pm 0.10$ & $78.50 \pm 0.67$& $90.97 \pm 0.11$  & $78.02 \pm 0.43$& $91.09 \pm 0.16$  & $77.32 \pm 0.46$ & $93.06 \pm 0.12$ & $84.15 \pm 0.35$\\
& KSE & $92.06 \pm 0.11$ & $78.83 \pm 0.47$& $92.50 \pm 0.08$  & $80.02 \pm 0.38$ & $92.31 \pm 0.19$ & $79.75 \pm 0.61$& $94.68 \pm 0.12$ & $86.01 \pm 0.42$ \\
& clustered\_SE & $90.08 \pm 0.08$& $77.04 \pm 0.25$ & $90.93 \pm 0.14$  & $78.39 \pm 0.35$ & $91.98 \pm 0.08$ & $81.38 \pm 0.21$ & $94.01 \pm 0.14$ & $84.00 \pm 0.38$\\
& $\alpha$\_clustered\_SE & $93.94 \pm 0.06$ & $84.91 \pm 0.23$& $94.48 \pm 0.08$ & $86.18 \pm 0.32$  & $94.57 \pm 0.09$ & $86.51 \pm 0.33$& $95.65 \pm 0.10$  & $88.77 \pm 0.24$ \\
& EigV & $93.85 \pm 0.07$ & $84.18 \pm 0.31$  & $94.60 \pm 0.08$& $86.18 \pm 0.41$  & $94.26 \pm 0.08$& $84.81 \pm 0.28$  & $95.45 \pm 0.10$ & $88.06 \pm 0.22$ \\
& LS & $86.16 \pm 0.70$ & $30.14^\dagger \pm 19.93$ & $85.57 \pm 0.75$& $0.00^\dagger \pm 0.00$  & $88.65 \pm 0.66$ & $58.28^\dagger \pm 4.08$ & $88.89 \pm 0.91$ & $60.32^\dagger \pm 2.49$\\
& \textbf{Ours} & $\mathbf{94.30}^* \pm \mathbf{0.09}$& $\mathbf{85.80}^* \pm \mathbf{0.56}$ & $\mathbf{94.82}^* \pm \mathbf{0.07}$  & $\mathbf{87.12}^* \pm \mathbf{0.43}$ & $\mathbf{94.78}^* \pm \mathbf{0.15}$ & $\mathbf{86.52}^* \pm \mathbf{0.64}$& $\mathbf{95.87}^* \pm \mathbf{0.11}$ & $\mathbf{89.80}^* \pm \mathbf{0.41}$ \\

\bottomrule
\end{tabular}
\end{adjustbox}
\label{tab: auroc of coqa and trivia qa(10 repeated exps)}
\end{table*}
\textbf{Effectiveness of our method.}
Table~\ref{tab: auroc of coqa and trivia qa(10 repeated exps)} reports AUROC and detection power at a fixed $10\%$ false-alarm rate across datasets and models (for HaluEval, we set $\theta=0.2$). We emphasize that \textbf{our approach serves as a universal evaluation framework, delivers robust performance across diverse models and datasets, and achieves the best results in most settings}.
Our method consistently achieves superior detection power in each evaluated dataset-model pair compared to all baselines. Regarding AUROC, our method improves upon the strongest baseline by at least $0.65\%$ on CoQA and $0.22\%$ on TriviaQA; on HaluEval, our performance is comparable to the best-performing baseline (either exceeding it by at least $0.77\%$ or trailing it by at most $0.16\%$).
Notably, relative to the worst-performing baseline, our method yields substantial gains in AUROC ranging from $8.5\%$ to $28.3\%$.


\begin{figure}[t]
    \centering
    
    \begin{subfigure}[b]{1.1\linewidth}
        \includegraphics[width=\linewidth]{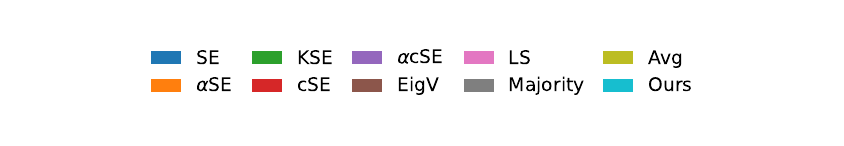}
    \end{subfigure}
    \vspace{-10mm} 

    \begin{subfigure}[b]{0.95\linewidth}
        \centering
        \includegraphics[width=\linewidth]{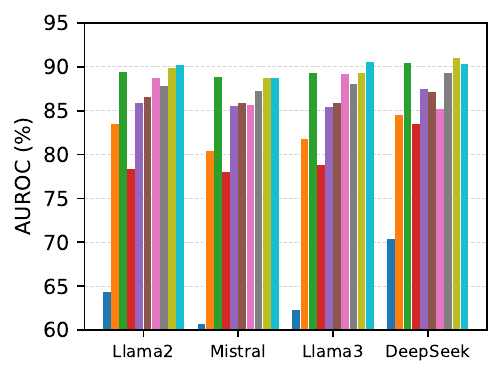}
    \end{subfigure}
    \vspace{-0.4cm}
    \begin{subfigure}[b]{0.95\linewidth}
        \centering
        \includegraphics[width=\linewidth]{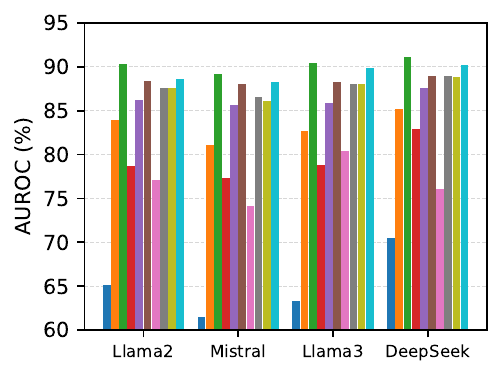}
    \end{subfigure}

    \caption{AUROC on the HaluEval dataset under Rouge-L annotation (top) and Llama annotation (bottom).}
    \label{fig: AUROC on HaluEval with diff annotations}
\end{figure}

We further study the effect of annotation protocols, comparing Rouge-L annotation with LLM-as-a-judge annotation.
For the latter, we use Llama-3.1-8B-Instruct to score consistency between the reference answer and the generated answer; we deem an answer correct if its consistency score exceeds $80$ out of $100$ (denoted as Llama annotation).
As shown in Figure~\ref{fig: AUROC on HaluEval with diff annotations}, the performance of individual baselines can shift under different annotations:
SE-based scores and the spectral eigenvalue score improve a little under Llama annotation, while lexical similarity degrades substantially because it is sensitive to surface overlap and aligns more naturally with Rouge-L.
Despite these shifts, even when some baselines degrade, our method remains robust, staying close to the best score and substantially improving over the weakest baseline.

Figure~\ref{fig: AUROC on HaluEval with diff annotations} also includes two simple aggregation strategies, majority voting and averaging.
Our aggregation outperforms them in most cases and remains competitive otherwise.
Notably, under Llama annotation, neither majority voting nor averaging remains close to the best baseline due to the influence of poorly performing scores, whereas our method consistently approaches the best score.
This highlights the more consistently robust behavior of our approach.

\begin{figure}
\centering
 \begin{adjustbox}{width=0.95\linewidth}
\includegraphics[]{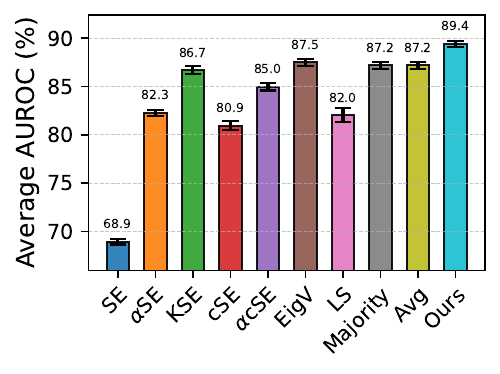}
\end{adjustbox}
\vspace{-0.4cm}
\caption{Average AUROC on all model-dataset pairs with Rouge-L annotation and Llama annotation.}
\label{fig: average AUROC}
\end{figure}

To further validate robustness, Figure~\ref{fig: average AUROC} aggregates results over all model--dataset pairs (with per-pair results reported in the appendix).
Our method achieves the highest average AUROC among all individual baselines and the two aggregation baselines, supporting its robustness across settings.


Because our approach makes no prior assumptions about datasets or models, a particular metric may outperform our method in specific cases.
However, such gains are typically small, and every baseline score deteriorates in some regimes.
For example, SE, $\alpha$SE, KSE, and EigV perform worse on GSM8K; clustered\_SE and $\alpha$\_clustered\_SE perform poorly on Qwen3; and LS drops substantially under Llama annotation, despite being among the strongest scores on GSM8K.
These inconsistencies suggest that individual scores are unreliable when the test-time prompt distribution is unknown.
In summary, our method effectively integrates the strengths of different detection signals, yielding robust and consistent performance across models and datasets.

\textbf{Ablation study.} 
As reasoning models become increasingly common, we additionally evaluate newer models such as Qwen3-4B. We observe that semantic-entropy variants that rely only on cluster-assignment frequencies can degrade for long-form generations: long answers often differ in meaning to some extent, which makes clustering  less informative and can weaken such scores. Because our method aggregates multiple scores, severely degraded baselines can in turn reduce overall performance. On CoQA, our method still outperforms all alternatives except EigV and LS.


We also study the effect of the calibration set size. In principle, increasing the calibration size reduces $\epsilon$ and may improve detection power. When we increase the calibration set from $1{,}000$ to $2{,}000$, the AUROC improvement is modest (up to $0.2\%$). Meanwhile, the standard deviation increases by more than $0.2\%$, likely because allocating more samples to calibration leaves fewer non-hallucinated examples for validating the false-alarm rate.


Next, we further enlarge the calibration set by merging CoQA and TriviaQA and sampling $3{,}000$ calibration prompts without hallucinated generations from the combined pool. This setting serves as a stress test for the realistic scenario where the test-time prompt distribution is unknown. Under this mixed calibration set, our performance changes only slightly, ranging from $-0.21\%$ to $0.27\%$ on CoQA and TriviaQA.

In many LLM tasks, the available validation/test splits are relatively small, making it challenging to obtain a calibration set large enough to fully meet theoretical desiderata. Our results suggest that very large calibration sets are not required for strong empirical performance, and that mixing calibration data across distributions does not materially degrade performance.

Finally, the sensitivity of ROUGE-L annotations is analyzed by varying the tolerance threshold $\theta\in\{0.1,0.2,0.3\}$ and the ROUGE-L similarity threshold $\tau\in\{0.2,0.3,0.4\}$. As shown in Table \ref{tab:sensitivity_theta_appendix} in the Appendix, increasing $\theta$ inherently increases task difficulty by introducing more noise into the non-hallucinated calibration set. While this leads to a general decline in AUROC across all detectors, our proposed aggregation method remains the top or second-best performer in all settings. Notably, while base signals like Spectral Eigenvalue and Clustered SE exhibit high sensitivity to $\theta$ (dropping over $8\%$ in AUROC on Mistral-7B), our framework maintains stability. This demonstrates that the multiple-testing procedure successfully leverages the most reliable signals available, even when other base detectors degrade. Furthermore, performance across all evaluated methods remained invariant to changes in $\tau$, indicating that the framework is not overly sensitive to the specific similarity threshold used for ground-truth labeling.

\section{Conclusions}
In this work, we reconceptualize the hallucination detection problem as a hypothesis testing problem and introduce a multiple-testing-inspired approach to integrate various hallucination detection methods. Our method provides a theoretical guarantee of false-alarm control while empirically improving AUROC across different evaluation metrics. 
Notably, whereas existing methods can be inconsistent across metrics, our approach achieves superior or comparable performance across a broad range of models and datasets, showing robust generalization without requiring assumptions about the distribution of user queries posed to large language models (LLMs). Our proposed method has significant implications for the reliability and trustworthiness of LLM-generated content, particularly in critical applications such as healthcare, where misinformation can have severe consequences. By providing a robust hallucination detection mechanism with controllable false-alarm rates and negligible additional overhead beyond computing existing scores,
our work supports safer deployment of LLM-based systems and reduces the risk that misleading or fabricated information is presented as fact.

\section{Limitations}

Despite its strengths, our method has a few limitations. Firstly, our evaluation relies on labeling prompts as inducing hallucinations or not; we currently use Rouge-L overlap or an LLM-based consistency judge. More capable LLM judges may better capture semantic equivalence under rephrasing and thus provide higher-quality ground-truth labels.

Secondly, as reasoning models show strong potential on complex tasks, further evaluation on such models, especially on more challenging settings such as complex mathematical reasoning or agentic problems, would be a valuable direction for future work.

Finally, our approach is aggregation-based and operates on a collection of base hallucination scores. In settings where one can propose multiple potentially reasonable candidate scores but it is unclear which will transfer best, this design is a feature: our procedure provides a principled way to combine them while controlling false alarms. That said, our method presumes access to at least one meaningful base score; for tasks where constructing such a score is itself challenging, score design remains a prerequisite and is complementary to our contribution.


\section*{Acknowledgment} 
This work was supported by the U.S. National Science Foundation (NSF) under grant 2106727.
The authors also acknowledge the Advanced Cyberinfrastructure Coordination Ecosystem: Services \& Support (ACCESS)  program~\cite{boerner2023access} for providing computing allocations and support through allocation CIS250917  (NSF grants \#2138259, \#2138286, \#2138307, \#2137603, and \#2138296).
\bibliography{ref}

@article{nikitin2024kernel,
  title={Kernel language entropy: Fine-grained uncertainty quantification for llms from semantic similarities},
  author={Nikitin, Alexander and Kossen, Jannik and Gal, Yarin and Marttinen, Pekka},
  journal={Advances in Neural Information Processing Systems},
  volume={37},
  pages={8901--8929},
  year={2024}
}

@article{yang2025qwen3,
  title={Qwen3 technical report},
  author={Yang, An and Li, Anfeng and Yang, Baosong and Zhang, Beichen and Hui, Binyuan and Zheng, Bo and Yu, Bowen and Gao, Chang and Huang, Chengen and Lv, Chenxu and others},
  journal={arXiv preprint arXiv:2505.09388},
  year={2025}
}

@article{yang2024qwen2,
  title={Qwen2. 5-math technical report: Toward mathematical expert model via self-improvement},
  author={Yang, An and Zhang, Beichen and Hui, Binyuan and Gao, Bofei and Yu, Bowen and Li, Chengpeng and Liu, Dayiheng and Tu, Jianhong and Zhou, Jingren and Lin, Junyang and others},
  journal={arXiv preprint arXiv:2409.12122},
  year={2024}
}

@article{li2023halueval,
  title={Halueval: A large-scale hallucination evaluation benchmark for large language models},
  author={Li, Junyi and Cheng, Xiaoxue and Zhao, Wayne Xin and Nie, Jian-Yun and Wen, Ji-Rong},
  journal={arXiv preprint arXiv:2305.11747},
  year={2023}
}

@article{benjamini2001,
 ISSN = {00905364},
 author = {Yoav Benjamini and Daniel Yekutieli},
 journal = {The Annals of Statistics},
 number = {4},
 pages = {1165--1188},
 publisher = {Institute of Mathematical Statistics},
 title = {The Control of the False Discovery Rate in Multiple Testing under Dependency},
 urldate = {2022-05-19},
 volume = {29},
 year = {2001}
}

@InProceedings{conformal,
  title = 	 {Conditional Validity of Inductive Conformal Predictors},
  author = 	 {Vovk, Vladimir},
  booktitle = 	 {Proceedings of the Asian Conference on Machine Learning},
  pages = 	 {475--490},
  year = 	 {2012},
  volume = 	 {25},
  series = 	 {Proceedings of Machine Learning Research},
  publisher =    {PMLR}
}

@article{ji2023survey,
  title={Survey of hallucination in natural language generation},
  author={Ji, Ziwei and Lee, Nayeon and Frieske, Rita and Yu, Tiezheng and Su, Dan and Xu, Yan and Ishii, Etsuko and Bang, Ye Jin and Madotto, Andrea and Fung, Pascale},
  journal={ACM Computing Surveys},
  volume={55},
  number={12},
  pages={1--38},
  year={2023},
  publisher={ACM New York, NY}
}

@article{lin2023generating,
  title={Generating with confidence: Uncertainty quantification for black-box large language models},
  author={Lin, Zhen and Trivedi, Shubhendu and Sun, Jimeng},
  journal={arXiv preprint arXiv:2305.19187},
  year={2023}
}

@article{magesh2023principled,
  title={Principled out-of-distribution detection via multiple testing},
  author={Magesh, Akshayaa and Veeravalli, Venugopal V and Roy, Anirban and Jha, Susmit},
  journal={Journal of Machine Learning Research},
  volume={24},
  number={378},
  pages={1--35},
  year={2023}
}

@article{kuhn2023semantic,
  title={Semantic uncertainty: Linguistic invariances for uncertainty estimation in natural language generation},
  author={Kuhn, Lorenz and Gal, Yarin and Farquhar, Sebastian},
  journal={arXiv preprint arXiv:2302.09664},
  year={2023}
}

@article{rateike2023weakly,
  title={Weakly supervised detection of hallucinations in llm activations},
  author={Rateike, Miriam and Cintas, Celia and Wamburu, John and Akumu, Tanya and Speakman, Skyler},
  journal={arXiv preprint arXiv:2312.02798},
  year={2023}
}

@article{kaur2024addressing,
  title={Addressing Uncertainty in LLMs to Enhance Reliability in Generative AI},
  author={Kaur, Ramneet and Samplawski, Colin and Cobb, Adam D and Roy, Anirban and Matejek, Brian and Acharya, Manoj and Elenius, Daniel and Berenbeim, Alexander M and Pavlik, John A and Bastian, Nathaniel D and others},
  journal={arXiv preprint arXiv:2411.02381},
  year={2024}
}

@article{fomicheva2020unsupervised,
  title={Unsupervised quality estimation for neural machine translation},
  author={Fomicheva, Marina and Sun, Shuo and Yankovskaya, Lisa and Blain, Fr{\'e}d{\'e}ric and Guzm{\'a}n, Francisco and Fishel, Mark and Aletras, Nikolaos and Chaudhary, Vishrav and Specia, Lucia},
  journal={Transactions of the Association for Computational Linguistics},
  volume={8},
  pages={539--555},
  year={2020},
  publisher={MIT Press One Rogers Street, Cambridge, MA 02142-1209, USA journals-info~…}
}

@article{grattafiori2024llama,
  title={The llama 3 herd of models},
  author={Grattafiori, Aaron and Dubey, Abhimanyu and Jauhri, Abhinav and Pandey, Abhinav and Kadian, Abhishek and Al-Dahle, Ahmad and Letman, Aiesha and Mathur, Akhil and Schelten, Alan and Vaughan, Alex and others},
  journal={arXiv preprint arXiv:2407.21783},
  year={2024}
}

@article{touvron2023llama,
  title={Llama 2: Open foundation and fine-tuned chat models},
  author={Touvron, Hugo and Martin, Louis and Stone, Kevin and Albert, Peter and Almahairi, Amjad and Babaei, Yasmine and Bashlykov, Nikolay and Batra, Soumya and Bhargava, Prajjwal and Bhosale, Shruti and others},
  journal={arXiv preprint arXiv:2307.09288},
  year={2023}
}

@misc{deepseekv2,
      title={DeepSeek-V2: A Strong, Economical, and Efficient Mixture-of-Experts Language Model}, 
      author={DeepSeek-AI},
      year={2024},
      eprint={2405.04434},
      archivePrefix={arXiv},
      primaryClass={cs.CL}
}

@misc{jiang2023mistral7b,
      title={Mistral 7B}, 
      author={Albert Q. Jiang and Alexandre Sablayrolles and Arthur Mensch and Chris Bamford and Devendra Singh Chaplot and Diego de las Casas and Florian Bressand and Gianna Lengyel and Guillaume Lample and Lucile Saulnier and Lélio Renard Lavaud and Marie-Anne Lachaux and Pierre Stock and Teven Le Scao and Thibaut Lavril and Thomas Wang and Timothée Lacroix and William El Sayed},
      year={2023},
      eprint={2310.06825},
      archivePrefix={arXiv},
      primaryClass={cs.CL},
      url={https://arxiv.org/abs/2310.06825}, 
}

@article{reddy2019coqa,
  title={Coqa: A conversational question answering challenge},
  author={Reddy, Siva and Chen, Danqi and Manning, Christopher D},
  journal={Transactions of the Association for Computational Linguistics},
  volume={7},
  pages={249--266},
  year={2019},
  publisher={MIT Press One Rogers Street, Cambridge, MA 02142-1209, USA journals-info~…}
}

@article{2017arXivtriviaqa,
       author = {{Joshi}, Mandar and {Choi}, Eunsol and {Weld},
                 Daniel and {Zettlemoyer}, Luke},
        title = "{triviaqa: A Large Scale Distantly Supervised Challenge Dataset for Reading Comprehension}",
      journal = {arXiv e-prints},
         year = 2017,
          eid = {arXiv:1705.03551},
        pages = {arXiv:1705.03551},
archivePrefix = {arXiv},
       eprint = {1705.03551},
}

@article{yao2023llm,
  title={Llm lies: Hallucinations are not bugs, but features as adversarial examples},
  author={Yao, Jia-Yu and Ning, Kun-Peng and Liu, Zhen-Hui and Ning, Mu-Nan and Liu, Yu-Yang and Yuan, Li},
  journal={arXiv preprint arXiv:2310.01469},
  year={2023}
}

@article{zhang2023language,
  title={How language model hallucinations can snowball},
  author={Zhang, Muru and Press, Ofir and Merrill, William and Liu, Alisa and Smith, Noah A},
  journal={arXiv preprint arXiv:2305.13534},
  year={2023}
}

@article{huang2025survey,
  title={A survey on hallucination in large language models: Principles, taxonomy, challenges, and open questions},
  author={Huang, Lei and Yu, Weijiang and Ma, Weitao and Zhong, Weihong and Feng, Zhangyin and Wang, Haotian and Chen, Qianglong and Peng, Weihua and Feng, Xiaocheng and Qin, Bing and others},
  journal={ACM Transactions on Information Systems},
  volume={43},
  number={2},
  pages={1--55},
  year={2025},
  publisher={ACM New York, NY}
}

@article{chern2023factool,
  title={FacTool: Factuality Detection in Generative AI--A Tool Augmented Framework for Multi-Task and Multi-Domain Scenarios},
  author={Chern, I and Chern, Steffi and Chen, Shiqi and Yuan, Weizhe and Feng, Kehua and Zhou, Chunting and He, Junxian and Neubig, Graham and Liu, Pengfei and others},
  journal={arXiv preprint arXiv:2307.13528},
  year={2023}
}

@article{huo2023retrieving,
  title={Retrieving supporting evidence for llms generated answers},
  author={Huo, Siqing and Arabzadeh, Negar and Clarke, Charles LA},
  journal={arXiv preprint arXiv:2306.13781},
  year={2023}
}

@article{zhou2020detecting,
  title={Detecting hallucinated content in conditional neural sequence generation},
  author={Zhou, Chunting and Neubig, Graham and Gu, Jiatao and Diab, Mona and Guzman, Paco and Zettlemoyer, Luke and Ghazvininejad, Marjan},
  journal={arXiv preprint arXiv:2011.02593},
  year={2020}
}

@article{luo2023chatgpt,
  title={Chatgpt as a factual inconsistency evaluator for text summarization},
  author={Luo, Zheheng and Xie, Qianqian and Ananiadou, Sophia},
  journal={arXiv preprint arXiv:2303.15621},
  year={2023}
}

@article{manakul2023selfcheckgpt,
  title={Selfcheckgpt: Zero-resource black-box hallucination detection for generative large language models},
  author={Manakul, Potsawee and Liusie, Adian and Gales, Mark JF},
  journal={arXiv preprint arXiv:2303.08896},
  year={2023}
}

@article{muhammed2025selfcheckagent,
  title={SelfCheckAgent: Zero-Resource Hallucination Detection in Generative Large Language Models},
  author={Muhammed, Diyana and Rabby, Gollam and Auer, S{\"o}ren},
  journal={arXiv preprint arXiv:2502.01812},
  year={2025}
}

@article{yang2025hallucination,
  title={Hallucination Detection in Large Language Models with Metamorphic Relations},
  author={Yang, Borui and Mamun, Md Afif Al and Zhang, Jie M and Uddin, Gias},
  journal={arXiv preprint arXiv:2502.15844},
  year={2025}
}

@article{varshney2023stitch,
  title={A stitch in time saves nine: Detecting and mitigating hallucinations of llms by validating low-confidence generation},
  author={Varshney, Neeraj and Yao, Wenlin and Zhang, Hongming and Chen, Jianshu and Yu, Dong},
  journal={arXiv preprint arXiv:2307.03987},
  year={2023}
}

@article{achiam2023gpt,
  title={Gpt-4 technical report},
  author={Achiam, Josh and Adler, Steven and Agarwal, Sandhini and Ahmad, Lama and Akkaya, Ilge and Aleman, Florencia Leoni and Almeida, Diogo and Altenschmidt, Janko and Altman, Sam and Anadkat, Shyamal and others},
  journal={arXiv preprint arXiv:2303.08774},
  year={2023}
}

@article{caruccio2024claude,
  title={Claude 2.0 large language model: Tackling a real-world classification problem with a new iterative prompt engineering approach},
  author={Caruccio, Loredana and Cirillo, Stefano and Polese, Giuseppe and Solimando, Giandomenico and Sundaramurthy, Shanmugam and Tortora, Genoveffa},
  journal={Intelligent Systems with Applications},
  volume={21},
  pages={200336},
  year={2024},
  publisher={Elsevier}
}

@article{team2024gemini,
  title={Gemini 1.5: Unlocking multimodal understanding across millions of tokens of context},
  author={Team, Gemini and Georgiev, Petko and Lei, Ving Ian and Burnell, Ryan and Bai, Libin and Gulati, Anmol and Tanzer, Garrett and Vincent, Damien and Pan, Zhufeng and Wang, Shibo and others},
  journal={arXiv preprint arXiv:2403.05530},
  year={2024}
}

@article{onoe2022entity,
  title={Entity cloze by date: What LMs know about unseen entities},
  author={Onoe, Yasumasa and Zhang, Michael JQ and Choi, Eunsol and Durrett, Greg},
  journal={arXiv preprint arXiv:2205.02832},
  year={2022}
}

@inproceedings{ladhak2023pre,
  title={When do pre-training biases propagate to downstream tasks? a case study in text summarization},
  author={Ladhak, Faisal and Durmus, Esin and Suzgun, Mirac and Zhang, Tianyi and Jurafsky, Dan and McKeown, Kathleen and Hashimoto, Tatsunori B},
  booktitle={Proceedings of the 17th Conference of the European Chapter of the Association for Computational Linguistics},
  pages={3206--3219},
  year={2023}
}

@article{sharma2023towards,
  title={Towards understanding sycophancy in language models},
  author={Sharma, Mrinank and Tong, Meg and Korbak, Tomasz and Duvenaud, David and Askell, Amanda and Bowman, Samuel R and Cheng, Newton and Durmus, Esin and Hatfield-Dodds, Zac and Johnston, Scott R and others},
  journal={arXiv preprint arXiv:2310.13548},
  year={2023}
}

@article{farquhar2024detecting,
  title={Detecting hallucinations in large language models using semantic entropy},
  author={Farquhar, Sebastian and Kossen, Jannik and Kuhn, Lorenz and Gal, Yarin},
  journal={Nature},
  volume={630},
  number={8017},
  pages={625--630},
  year={2024},
  publisher={Nature Publishing Group UK London}
}

@article{raffel2020exploring,
  title={Exploring the limits of transfer learning with a unified text-to-text transformer},
  author={Raffel, Colin and Shazeer, Noam and Roberts, Adam and Lee, Katherine and Narang, Sharan and Matena, Michael and Zhou, Yanqi and Li, Wei and Liu, Peter J},
  journal={Journal of machine learning research},
  volume={21},
  number={140},
  pages={1--67},
  year={2020}
}

@article{singhal2025toward,
  title={Toward expert-level medical question answering with large language models},
  author={Singhal, Karan and Tu, Tao and Gottweis, Juraj and Sayres, Rory and Wulczyn, Ellery and Amin, Mohamed and Hou, Le and Clark, Kevin and Pfohl, Stephen R and Cole-Lewis, Heather and others},
  journal={Nature Medicine},
  pages={1--8},
  year={2025},
  publisher={Nature Publishing Group US New York}
}

@article{keskar2019ctrl,
  title={Ctrl: A conditional transformer language model for controllable generation},
  author={Keskar, Nitish Shirish and McCann, Bryan and Varshney, Lav R and Xiong, Caiming and Socher, Richard},
  journal={arXiv preprint arXiv:1909.05858},
  year={2019}
}

@article{zhang2024benchmarking,
  title={Benchmarking large language models for news summarization},
  author={Zhang, Tianyi and Ladhak, Faisal and Durmus, Esin and Liang, Percy and McKeown, Kathleen and Hashimoto, Tatsunori B},
  journal={Transactions of the Association for Computational Linguistics},
  volume={12},
  pages={39--57},
  year={2024},
  publisher={MIT Press One Broadway, 12th Floor, Cambridge, Massachusetts 02142, USA~…}
}

@article{cobbe2021gsm8k,
  title={Training Verifiers to Solve Math Word Problems},
  author={Cobbe, Karl and Kosaraju, Vineet and Bavarian, Mohammad and Chen, Mark and Jun, Heewoo and Kaiser, Lukasz and Plappert, Matthias and Tworek, Jerry and Hilton, Jacob and Nakano, Reiichiro and Hesse, Christopher and Schulman, John},
  journal={arXiv preprint arXiv:2110.14168},
  year={2021}
}

@article{yang2024qwen25mathtechnicalreportmathematical,
  title={Qwen2.5-Math Technical Report: Toward Mathematical Expert Model via Self-Improvement}, 
  author={An Yang and Beichen Zhang and Binyuan Hui and Bofei Gao and Bowen Yu and Chengpeng Li and Dayiheng Liu and Jianhong Tu and Jingren Zhou and Junyang Lin and Keming Lu and Mingfeng Xue and Runji Lin and Tianyu Liu and Xingzhang Ren and Zhenru Zhang},
  journal={arXiv preprint arXiv:2409.12122},
  year={2024}
}

@incollection{boerner2023access,
  title={Access: Advancing innovation: Nsf’s advanced cyberinfrastructure coordination ecosystem: Services \& support},
  author={Boerner, Timothy J and Deems, Stephen and Furlani, Thomas R and Knuth, Shelley L and Towns, John},
  booktitle={Practice and experience in advanced research computing 2023: Computing for the common good},
  pages={173--176},
  year={2023}
}

@article{lee2018simple,
  title={A simple unified framework for detecting out-of-distribution samples and adversarial attacks},
  author={Lee, Kimin and Lee, Kibok and Lee, Honglak and Shin, Jinwoo},
  journal={Advances in neural information processing systems},
  volume={31},
  year={2018}
}

@article{de2025mysteries,
  title={Mysteries of the Deep: Role of Intermediate Representations in Out of Distribution Detection},
  author={De la Jara, Ignacio Meza and Rodriguez-Opazo, Cristian and Teney, Damien and Ranasinghe, Damith and Abbasnejad, Ehsan},
  journal={arXiv preprint arXiv:2510.05782},
  year={2025}
}

\ifappendix
\appendix
\newpage
\section{Experiment Details}
\label{appendix: experiment details}
Table~\ref{tab: models} shows the specific models used in our experiments. For each model, we generate $20$ samples using pure sampling from the logits, with $\mathrm{top\_k}=0$, $\mathrm{top\_p}=1.0$, and $\mathrm{temperature}=1.0$. The hyperparameter $\alpha$ in Alpha Semantic Entropy score is fixed at $0.5$. All experiments are conducted on a single L40S GPU, except for LLaMA-2-13B, which is evaluated using two L40S GPUs. Sampling generations and computing scores for each test take approximately three days, but the process can be accelerated through parallel computing.


The computational overhead is not expensive in our experiments. Retrieving token probabilities requires most of the time (several hours), and after that, different versions of semantic entropy computations are finished within one hour separately, spectral eigenvalue and lexical similarity could be finished within one hour separately too. Our multiple testing framework is finished within $0.1$ second then.

We adopt the QA split of HaluEval with $10,000$ questions, the validation split of TriviaQA with $9,960$ questions, the development split of CoQA with $7,983$ questions, 
and the test split of GSM8K with $1319$ questions. 
Since GSM8K is harder for LLMs to answer, $\theta$ was chosen to be $0.3$, and the calibration dataset size is $500$.


\begin{table}[h]
\caption{Models tested in our experiments and their Hugging Face identifiers}
\label{tab: models}
\centering
\begin{adjustbox}{width=\linewidth}
\begin{tabular}{|c|c|}
\hline Model & Hugging Face Identifier\\
\hline
Llama-2-13B & meta-llama/Llama-2-13b-hf\\
\hline
Mistral-7B & mistralai/Mistral-7B-v0.1\\
\hline
Llama-3.1-8B & meta-llama/Llama-3.1-8B \\
\hline
DeepSeek-v2-Lite & deepseek-ai/DeepSeek-V2-Lite\\
\hline
Qwen2.5-Math-1.5B-Instruct & Qwen/Qwen2.5-Math-1.5B-Instruct \\
\hline
Llama-3.2-3B-Instruct & meta-llama/Llama-3.2-3B-Instruct\\
\hline
Qwen3-4B &Qwen/Qwen3-4B\\
\hline 
\end{tabular}
\end{adjustbox}

\end{table}

Table \ref{tab:hallucination_rates} reports the hallucination proportions ($\text{\# hallucinated samples}/\text{\# Total samples}$) derived from the experimental splits for each model-dataset pair. These statistics reveal that hallucination prevalence varies significantly across model architectures and annotation regimes. In many instances, such as within the HaluEval and TriviaQA benchmarks, hallucination rates exceed $70\%$, creating  high-noise environments. The consistency of our results across these diverse proportions underscores the robustness of the proposed framework.

\begin{table}[h]
\caption{Hallucination proportions (\%) across benchmarks.}
\centering
\begin{adjustbox}{width=\linewidth}
\begin{tabular}{llcc}
\hline \hline 
\multirow{2}{*}{Dataset} & \multirow{2}{*}{Model} & \multicolumn{2}{c}{Hallucination Rate (\%)} \\
\cline{3-4}
& & (Rouge-L Annotation) & (Llama Annotation) \\
\hline \hline
\multirow{4}{*}{HaluEval} 
& Llama-3.1-8B   & 70.89 & 81.01 \\
& DeepSeek-V2    & 71.97 & 80.61 \\
& Llama-2-13B    & 79.11 & 85.97 \\
& Mistral-7B     & 70.53 & 80.04 \\
\hline
\multirow{4}{*}{TriviaQA} 
& Llama-3.1-8B   & 74.75 & 78.73 \\
& Llama-2-13B    & 72.56 & 76.32 \\
& Mistral-7B     & 73.53 & 76.46 \\
& DeepSeek-V2    & 69.20 & 72.66 \\
\hline
\multirow{5}{*}{CoQA} 
& Llama-3.1-8B   & 63.47 & 83.55 \\
& Llama-2-13B    & 64.90 & 83.70 \\
& DeepSeek-V2    & 72.59 & 88.36 \\
& Mistral-7B     & 68.53 & 86.45 \\
& Qwen3-4B       & 85.43 & 87.32 \\
\hline
\multirow{2}{*}{GSM8K} 
& Qwen2.5-Math   & 77.71 & 66.26 \\
& Llama-3.2-3B   & 55.88 & 47.61 \\
\hline \hline
\end{tabular}
\end{adjustbox}
\label{tab:hallucination_rates}
\end{table}


\section{Additional Results}
\label{appendix: additional_result}

\begin{table}[t]
\caption{Detection power (\%) at 10\% false alarm rate across different models and datasets with a calibration data size of 3,000. $*$ and $\dagger$ indicate the best and the worst detection power in each case.}
\centering
\begin{adjustbox}{width=\linewidth}
\begin{tabular}{llcccc}
\hline \hline Dataset & Method & Llama-2-13B & Mistral-7B & Llama-3.1-8B & DeepSeek-v2-Lite \\

\hline \hline \multirow{8}{*}{ CoQA } 
& SE & $38.82^\dagger \pm 0.71$ & $36.65^\dagger \pm 0.66$ & $37.42^\dagger \pm 0.64$ & $41.86^\dagger \pm 0.56$ \\
& $\alpha$SE & $65.54 \pm 1.05$ & $65.45 \pm 1.41$ & $62.19 \pm 1.10$ & $66.65 \pm 0.88$ \\
& KSE & $70.46 \pm 0.92$ & $68.06 \pm 1.23$ & $67.44 \pm 0.93$ & $69.94 \pm 0.84$ \\
& clustered\_SE & $57.21 \pm 1.29$ & $58.50 \pm 1.42$ & $58.52 \pm 0.98$ & $60.79 \pm 1.06$ \\
& $\alpha$\_clustered\_SE & $72.79 \pm 0.71$ & $73.13 \pm 0.77$ & $71.65 \pm 0.89$ & $74.64 \pm 0.52$ \\
& EigV & $71.80 \pm 1.01$ & $73.42 \pm 0.80$ & $72.20 \pm 1.47$ & $74.01 \pm 1.30$ \\
& LS & $64.84 \pm 4.34$ & $67.41 \pm 1.20$ & $62.95 \pm 1.91$ & $70.32 \pm 1.27$ \\
& \textbf{Ours} & $\mathbf{74.48}^* \pm \mathbf{0.37}$ & $\mathbf{74.03}^* \pm \mathbf{1.18}$ & $\mathbf{73.55}^* \pm \mathbf{1.27}$ & $\mathbf{76.36}^* \pm \mathbf{1.22}$ \\

\hline \hline \multirow{8}{*}{ TriviaQA } 
& SE & $67.99 \pm 0.44$ & $66.52 \pm 0.48$ & $69.00 \pm 0.71$ & $75.30 \pm 0.26$ \\
& $\alpha$SE & $78.43 \pm 0.80$ & $77.55 \pm 0.50$ & $77.38 \pm 0.83$ & $83.92 \pm 0.50$ \\
& KSE & $78.71 \pm 0.58$ & $79.78 \pm 0.50$ & $79.71 \pm 0.91$ & $85.67 \pm 0.61$ \\
& clustered\_SE & $76.98 \pm 0.19$ & $78.27 \pm 0.70$ & $81.25 \pm 0.71$ & $84.04 \pm 0.60$ \\
& $\alpha$\_clustered\_SE & $84.84 \pm 0.37$ & $86.05 \pm 0.56$ & $86.47 \pm 0.64$ & $88.60 \pm 0.35$ \\
& EigV & $83.95 \pm 0.29$ & $86.01 \pm 0.41$ & $84.69 \pm 0.64$ & $87.92 \pm 0.36$ \\
& LS & $10.44^\dagger \pm 13.99$ & $0.00^\dagger \pm 0.00$ & $57.72^\dagger \pm 3.76$ & $60.66^\dagger \pm 6.73$ \\
& \textbf{Ours} & $\mathbf{86.52}^* \pm \mathbf{0.34}$ & $\mathbf{87.07}^* \pm \mathbf{0.71}$ & $\mathbf{87.82}^* \pm \mathbf{0.70}$ & $\mathbf{89.93}^* \pm \mathbf{0.25}$ \\

\hline \hline \multirow{8}{*}{ \parbox{1.5cm}{CoQA \\ +TriviaQA} } 
& SE & $54.96 \pm 0.55$ & $52.70 \pm 0.36$ & $56.64^\dagger \pm 0.41$ & $58.79^\dagger \pm 0.18$ \\
& $\alpha$SE & $73.03 \pm 0.54$ & $72.65 \pm 0.59$ & $70.72 \pm 0.52$ & $76.73 \pm 0.62$ \\
& KSE & $74.93 \pm 0.38$ & $74.60 \pm 0.77$ & $74.38 \pm 0.46$ & $79.07 \pm 0.76$ \\
& clustered\_SE & $67.20 \pm 0.24$ & $68.27 \pm 0.46$ & $68.99 \pm 0.31$ & $73.15 \pm 0.38$ \\
& $\alpha$\_clustered\_SE & $79.67 \pm 0.30$ & $80.67 \pm 0.52$ & $80.42 \pm 0.43$ & $82.83 \pm 0.36$ \\
& EigV & $78.82 \pm 0.30$ & $80.78 \pm 0.46$ & $78.99 \pm 0.51$ & $82.18 \pm 0.64$ \\
& LS & $50.14^\dagger \pm 1.68$ & $51.85^\dagger \pm 2.87$ & $60.58 \pm 2.71$ & $59.93 \pm 2.67$ \\
& \textbf{Ours} & $\mathbf{81.48}^* \pm \mathbf{0.34}$ & $\mathbf{81.81}^* \pm \mathbf{0.55}$ & $\mathbf{81.81}^* \pm \mathbf{0.62}$ & $\mathbf{84.06}^* \pm \mathbf{0.49}$ \\

\hline \hline
\end{tabular}
\end{adjustbox}
\label{tab: detection power of coqa and trivia qa (10 repeated exps, 3000 data)}
\end{table}

\begin{table}[t]
\caption{AUROC (\%) across different models and datasets with a calibration data size of 3,000. $*$ and $\dagger$ indicate the best and worst AUROC in each case.}
\centering
\begin{adjustbox}{width=\linewidth}
\begin{tabular}{llcccc}
\hline \hline Dataset & Method & Llama-2-13B & Mistral-7B & Llama-3.1-8B & DeepSeek-v2-Lite \\

\hline \hline \multirow{8}{*}{ CoQA } 
& SE & $66.76^\dagger \pm 0.27$ & $65.46^\dagger \pm 0.38$ & $65.55^\dagger \pm 0.31$ & $68.99^\dagger \pm 0.31$ \\
& $\alpha$SE & $85.21 \pm 0.20$ & $85.28 \pm 0.38$ & $84.29 \pm 0.39$ & $86.48 \pm 0.21$ \\
& KSE & $88.77 \pm 0.30$ & $88.53 \pm 0.26$ & $87.85 \pm 0.28$ & $89.03 \pm 0.27$ \\
& clustered\_SE & $85.68 \pm 0.25$ & $85.65 \pm 0.30$ & $85.41 \pm 0.37$ & $86.33 \pm 0.28$ \\
& $\alpha$\_clustered\_SE & $89.81 \pm 0.12$ & $90.08 \pm 0.20$ & $89.56 \pm 0.26$ & $90.73 \pm 0.23$ \\
& EigV & $89.98 \pm 0.25$ & $90.36 \pm 0.29$ & $89.85 \pm 0.25$ & $91.00 \pm 0.31$ \\
& LS & $88.49 \pm 0.57$ & $89.17 \pm 0.40$ & $88.01 \pm 0.48$ & $89.72 \pm 0.49$ \\
& \textbf{Ours} & $\mathbf{90.82}^* \pm \mathbf{0.15}$ & $\mathbf{91.02}^* \pm \mathbf{0.20}$ & $\mathbf{90.52}^* \pm \mathbf{0.31}$ & $\mathbf{91.57}^* \pm \mathbf{0.25}$ \\

\hline \hline \multirow{8}{*}{ TriviaQA }
& SE & $82.57^\dagger \pm 0.21$ & $82.00^\dagger \pm 0.20$ & $84.77^\dagger \pm 0.36$ & $87.31^\dagger \pm 0.21$ \\
& $\alpha$SE & $90.68 \pm 0.17$ & $90.91 \pm 0.21$ & $91.16 \pm 0.26$ & $92.99 \pm 0.24$ \\
& KSE & $92.09 \pm 0.16$ & $92.46 \pm 0.17$ & $92.39 \pm 0.28$ & $94.63 \pm 0.28$ \\
& clustered\_SE & $90.06 \pm 0.08$ & $90.87 \pm 0.24$ & $92.02 \pm 0.26$ & $94.01 \pm 0.19$ \\
& $\alpha$\_clustered\_SE & $93.85 \pm 0.07$ & $94.46 \pm 0.18$ & $94.61 \pm 0.21$ & $95.62 \pm 0.15$ \\
& EigV & $93.77 \pm 0.09$ & $94.60 \pm 0.14$ & $94.28 \pm 0.19$ & $95.42 \pm 0.19$ \\
& LS & $85.11 \pm 0.65$ & $85.29 \pm 1.09$ & $88.40 \pm 0.80$ & $89.20 \pm 0.84$ \\
& \textbf{Ours} & $\mathbf{94.44}^* \pm \mathbf{0.10}$ & $\mathbf{94.81}^* \pm \mathbf{0.13}$ & $\mathbf{95.05}^* \pm \mathbf{0.22}$ & $\mathbf{95.94}^* \pm \mathbf{0.18}$ \\

\hline \hline \multirow{8}{*}{\parbox{1.5cm}{TriviaQA \\ +CoQA} } 
& SE & $75.35^\dagger \pm 0.18$ & $74.12^\dagger \pm 0.19$ & $76.45^\dagger \pm 0.26$ & $78.00^\dagger \pm 0.19$ \\
& $\alpha$SE & $88.28 \pm 0.18$ & $88.37 \pm 0.15$ & $88.26 \pm 0.22$ & $90.09 \pm 0.24$ \\
& KSE & $90.62 \pm 0.17$ & $90.72 \pm 0.17$ & $90.42 \pm 0.19$ & $92.39 \pm 0.27$ \\
& clustered\_SE & $87.56 \pm 0.09$ & $88.04 \pm 0.16$ & $88.50 \pm 0.24$ & $90.39 \pm 0.18$ \\
& $\alpha$\_clustered\_SE & $92.00 \pm 0.07$ & $92.48 \pm 0.14$ & $92.40 \pm 0.17$ & $93.63 \pm 0.14$ \\
& EigV & $92.11 \pm 0.13$ & $92.77 \pm 0.17$ & $92.40 \pm 0.17$ & $93.64 \pm 0.19$ \\
& LS & $86.40 \pm 0.60$ & $86.32 \pm 0.48$ & $88.35 \pm 0.39$ & $88.79 \pm 0.49$ \\
& \textbf{Ours} & $\mathbf{92.74}^* \pm \mathbf{0.11}$ & $\mathbf{93.07}^* \pm \mathbf{0.11}$ & $\mathbf{92.98}^* \pm \mathbf{0.17}$ & $\mathbf{94.01}^* \pm \mathbf{0.17}$ \\

\hline \hline
\end{tabular}
\end{adjustbox}
\label{tab: auroc of coqa and trivia qa(10 repeated exps, 3000 data)}
\end{table}

\begin{table}[t]
\caption{Detection power (\%) at 10\% false alarm rate across different models and datasets with a calibration data size of 2,000. $*$ and $\dagger$ indicate the best and the worst detection power in each case.}
\centering
\begin{adjustbox}{width=\linewidth}
\begin{tabular}{llcccc}
\hline \hline Dataset & Method & Llama-2-13B & Mistral-7B & Llama-3.1-8B & DeepSeek-v2-Lite \\

\hline \hline \multirow{8}{*}{ CoQA } 
& SE & $38.69^\dagger \pm 0.81$ & $37.55^\dagger \pm 1.44$ & $37.11^\dagger \pm 0.93$ & $42.73^\dagger \pm 2.12$ \\
& $\alpha$SE & $65.97 \pm 1.63$ & $65.11 \pm 2.79$ & $62.30 \pm 0.62$ & $66.28 \pm 4.25$ \\
& KSE & $70.67 \pm 1.28$ & $68.89 \pm 1.34$ & $67.27 \pm 1.02$ & $70.07 \pm 3.20$ \\
& clustered\_SE & $56.96 \pm 2.20$ & $59.03 \pm 2.58$ & $58.63 \pm 1.01$ & $59.31 \pm 3.31$ \\
& $\alpha$\_clustered\_SE & $72.57 \pm 1.35$ & $73.64 \pm 1.69$ & $71.47 \pm 1.09$ & $74.32 \pm 2.04$ \\
& EigV & $72.54 \pm 1.26$ & $74.34 \pm 1.49$ & $72.01 \pm 1.72$ & $72.77 \pm 3.01$ \\
& LS & $62.34 \pm 5.29$ & $68.91 \pm 2.71$ & $60.91 \pm 3.58$ & $65.05 \pm 8.28$ \\
& \textbf{Ours} & $\mathbf{74.72}^* \pm \mathbf{1.28}$ & $\mathbf{76.53}^* \pm \mathbf{2.70}$ & $\mathbf{74.69}^* \pm \mathbf{1.01}$ & $\mathbf{75.81}^* \pm \mathbf{2.47}$ \\

\hline \hline \multirow{8}{*}{ TriviaQA } 
& SE & $67.82 \pm 0.68$ & $67.12 \pm 0.82$ & $68.66 \pm 1.63$ & $75.48 \pm 0.55$ \\
& $\alpha$SE & $78.44 \pm 1.09$ & $77.98 \pm 1.19$ & $77.46 \pm 1.46$ & $84.16 \pm 0.55$ \\
& KSE & $78.84 \pm 1.39$ & $80.22 \pm 0.90$ & $79.65 \pm 1.16$ & $85.87 \pm 0.73$ \\
& clustered\_SE & $76.78 \pm 1.26$ & $78.40 \pm 0.86$ & $81.39 \pm 1.26$ & $83.91 \pm 0.48$ \\
& $\alpha$\_clustered\_SE & $84.93 \pm 0.77$ & $86.21 \pm 1.18$ & $86.33 \pm 1.07$ & $88.75 \pm 0.39$ \\
& EigV & $84.22 \pm 0.74$ & $86.30 \pm 1.00$ & $84.90 \pm 1.07$ & $88.11 \pm 0.49$ \\
& LS & $34.19^\dagger \pm 23.91$ & $0.00^\dagger \pm 0.00$ & $54.87^\dagger \pm 7.37$ & $58.38^\dagger \pm 6.66$ \\
& \textbf{Ours} & $\mathbf{85.79}^* \pm \mathbf{0.81}$ & $\mathbf{87.28}^* \pm \mathbf{1.01}$ & $\mathbf{86.52}^* \pm \mathbf{1.17}$ & $\mathbf{89.81}^* \pm \mathbf{0.60}$ \\

\hline \hline
\end{tabular}
\end{adjustbox}
\label{tab: detection power of coqa and trivia qa (10 repeated exps, 2000 data)}
\end{table}

\begin{table}[h]
\caption{AUROC (\%) across different models and datasets with a calibration data size of 2,000. $*$ and $\dagger$ indicate the best and worst AUROC in each case.}
\centering
\begin{adjustbox}{width=\linewidth}
\begin{tabular}{llcccc}
\hline \hline Dataset & Method & Llama-2-13B & Mistral-7B & Llama-3.1-8B & DeepSeek-v2-Lite \\

\hline \hline \multirow{8}{*}{ CoQA }
& SE & $66.59^\dagger \pm 0.49$ & $65.76^\dagger \pm 0.61$ & $65.35^\dagger \pm 0.31$ & $69.38^\dagger \pm 1.29$ \\
& $\alpha$SE & $85.38 \pm 0.40$ & $85.38 \pm 0.57$ & $84.17 \pm 0.33$ & $86.82 \pm 0.64$ \\
& KSE & $88.86 \pm 0.27$ & $88.53 \pm 0.55$ & $87.86 \pm 0.26$ & $89.46 \pm 0.67$ \\
& clustered\_SE & $85.55 \pm 0.70$ & $85.64 \pm 0.75$ & $85.25 \pm 0.27$ & $86.46 \pm 1.16$ \\
& $\alpha$\_clustered\_SE & $89.85 \pm 0.40$ & $90.09 \pm 0.50$ & $89.41 \pm 0.31$ & $90.94 \pm 0.51$ \\
& EigV & $90.10 \pm 0.30$ & $90.43 \pm 0.59$ & $89.80 \pm 0.35$ & $91.28 \pm 0.69$ \\
& LS & $88.16 \pm 0.77$ & $89.54 \pm 0.80$ & $87.87 \pm 0.78$ & $89.46 \pm 1.41$ \\
& \textbf{Ours} & $\mathbf{90.90}^* \pm \mathbf{0.36}$ & $\mathbf{91.44}^* \pm \mathbf{0.60}$ & $\mathbf{90.54}^* \pm \mathbf{0.26}$ & $\mathbf{91.70}^* \pm \mathbf{0.51}$ \\

\hline \hline \multirow{8}{*}{ TriviaQA }
& SE & $82.44^\dagger \pm 0.45$ & $82.19^\dagger \pm 0.50$ & $84.77^\dagger \pm 0.60$ & $87.36^\dagger \pm 0.22$ \\
& $\alpha$SE & $90.72 \pm 0.29$ & $91.03 \pm 0.48$ & $91.24 \pm 0.46$ & $93.08 \pm 0.20$ \\
& KSE & $92.06 \pm 0.26$ & $92.68 \pm 0.44$ & $92.30 \pm 0.40$ & $94.73 \pm 0.22$ \\
& clustered\_SE & $90.02 \pm 0.30$ & $90.98 \pm 0.43$ & $92.06 \pm 0.53$ & $93.99 \pm 0.21$ \\
& $\alpha$\_clustered\_SE & $93.94 \pm 0.22$ & $94.56 \pm 0.35$ & $94.65 \pm 0.40$ & $95.66 \pm 0.19$ \\
& EigV & $93.80 \pm 0.21$ & $94.72 \pm 0.33$ & $94.30 \pm 0.43$ & $95.49 \pm 0.16$ \\
& LS & $86.43 \pm 0.55$ & $85.40 \pm 1.29$ & $88.26 \pm 0.93$ & $89.31 \pm 0.56$ \\
& \textbf{Ours} & $\mathbf{94.29}^* \pm \mathbf{0.21}$ & $\mathbf{94.90}^* \pm \mathbf{0.31}$ & $\mathbf{94.81}^* \pm \mathbf{0.40}$ & $\mathbf{95.92}^* \pm \mathbf{0.18}$ \\

\hline \hline
\end{tabular}
\end{adjustbox}
\label{tab: auroc of coqa and trivia qa(10 repeated exps, 2000 data)}
\end{table}

\paragraph{Calibration Dataset Expansion.} Table~\ref{tab: detection power of coqa and trivia qa (10 repeated exps, 3000 data)} and Table~\ref{tab: auroc of coqa and trivia qa(10 repeated exps, 3000 data)} correspond to constructing a mixed calibration set by pooling non-hallucinated prompts from TriviaQA and CoQA. As noted in the main text, AUROC changes only slightly (within $0.3\%$), and detection power also varies modestly (within $1.8\%$).
Table~\ref{tab: detection power of coqa and trivia qa (10 repeated exps, 2000 data)} and Table~\ref{tab: auroc of coqa and trivia qa(10 repeated exps, 2000 data)} consider a second strategy, where we sample $2{,}000$ non-hallucinated prompts and use the remaining non-hallucinated prompts to estimate the false-alarm rate. Performance remains comparable to the default setting with $|\mathcal C|=1{,}000$. The standard deviation is larger, likely because the number of held-out non-hallucinated examples used to estimate the false-alarm rate is smaller in each case (ranging from $188$ to $1{,}068$). Overall, these results further support the robustness of our method, which remains the top-performing approach across all settings.

\paragraph{Results on math datasets.}Because math questions are more challenging, we focus on instruction-tuned models, Qwen2.5-Math-1.5B-Instruct~\citep{yang2024qwen25mathtechnicalreportmathematical} and LLaMA-3.2-3B-Instruct~\citep{grattafiori2024llama}, using chain-of-thought prompting with the chat-style template.
Because our evaluation setting aims not to assume any particular prompt type, we do not post-process generations (e.g., extracting the final numeric answer), as is common in prior work on math benchmarks. Instead, we keep the full generations, consistent with our other QA datasets. For annotation, however, we extract the final numeric answer from each sampled generation when it exists.


As shown in Table~\ref{tab: detection power of math} and Table~\ref{tab: auroc of math}, our method remains robust. Lexical similarity is the only baseline that achieves consistently strong performance, while several others degrade substantially (some even perform worse than random guessing). In the presence of these weak baselines, our method attains the second-best overall performance.
These differences arise because, across samples, math solutions often differ in their intermediate reasoning, rephrasings, or decomposition of the problem, which can distort semantic-similarity signals and degrade semantic-entropy variants as well as spectral scores. In contrast, lexical similarity primarily captures surface overlap, and thus better reflects whether two generations are following a similar solution trajectory rather than diverging into irrelevant content or generic refusals. We also report majority voting and averaging; our method substantially improves over these  aggregation strategies in most settings.

\begin{table}[h]
\caption{Detection power (\%) at 10\% false alarm rate across different models on GSM8K.  $*$ and $\dagger$ indicate the best and the worst detection power in each case.}
\centering
\begin{adjustbox}{width=\linewidth}
\begin{tabular}{llcc}
\hline \hline Dataset & Method & Qwen2.5-Math-1.5B-Instruct & Llama-3.2-3B-Instruct \\
\hline \hline \multirow{10}{*}{ GSM8K } 
& SE & $0.21^\dagger \pm 0.00$ & $0.77^\dagger \pm 0.00$ \\
& $\alpha$SE & $5.05 \pm 0.00$ & $3.23 \pm 0.00$ \\
& KSE & $8.84 \pm 1.43$ & $7.45 \pm 1.54$ \\
& clustered\_SE & $16.59 \pm 2.82$ & $28.12 \pm 2.52$ \\
& $\alpha$\_clustered\_SE & $12.75 \pm 0.64$ & $23.40 \pm 3.92$ \\
& EigV & $9.20 \pm 1.93$ & $6.02 \pm 1.72$ \\
& LS & $32.66^* \pm 2.73$ & $63.91^* \pm 3.94$ \\
& Majority Voting & $12.99 \pm 3.47$ & $16.05 \pm 2.92$ \\
& Averaging & $11.35 \pm 2.56$ & $35.26 \pm 4.69$ \\
& \textbf{Ours} & $\mathbf{18.19} \pm \mathbf{3.20}$ & $\mathbf{44.23} \pm \mathbf{4.83}$ \\

\hline \hline
\end{tabular}
\end{adjustbox}
\label{tab: detection power of math}
\end{table}

\begin{table}[h]
\caption{AUROC (\%) across different models on GSM8K. 
$*$ and $\dagger$ indicate the best and worst AUROC in each case.}
\centering
\begin{adjustbox}{width=\linewidth}
\begin{tabular}{llcc}
\hline \hline Dataset & Method & Qwen2.5-Math-1.5B-Instruct & Llama-3.2-3B-Instruct \\
\hline \hline \multirow{10}{*}{ GSM8K } 
& SE & $49.54 \pm 0.17$ & $50.21 \pm 0.21$ \\
& $\alpha$SE & $48.87 \pm 0.61$ & $49.47 \pm 0.75$ \\
& KSE & $47.99^\dagger \pm 1.45$ & $46.34 \pm 1.90$ \\
& clustered\_SE & $61.24 \pm 1.31$ & $63.61 \pm 1.77$ \\
& $\alpha$\_clustered\_SE & $60.00 \pm 1.61$ & $62.34 \pm 2.01$ \\
& EigV & $49.46 \pm 1.47$ & $43.57^\dagger \pm 1.92$ \\
& LS & $75.57^* \pm 1.06$ & $83.79^* \pm 0.75$ \\
& Majority Voting & $54.26 \pm 1.35$ & $54.33 \pm 1.84$ \\
& Averaging & $64.70 \pm 1.57$ & $71.92 \pm 1.50$ \\
& \textbf{Ours} & $\mathbf{62.24} \pm \mathbf{1.39}$ & $\mathbf{74.10} \pm \mathbf{1.58}$ \\

\hline \hline
\end{tabular}
\end{adjustbox}
\label{tab: auroc of math}
\end{table}

\begin{table}[h]
\caption{Detection power (\%) at 10\% false alarm rate and AUROC (\%) on Qwen3-4B and CoQA. $*$ and $\dagger$ indicate the best and worst performance in each case.}
\centering
\begin{adjustbox}{width=\linewidth}
\begin{tabular}{lcc}
\hline \hline Method & AUROC & Detection Power \\
\hline \hline
SE & $51.88 \pm 0.40$ & $9.15 \pm 0.00$ \\
$\alpha$SE & $61.51 \pm 0.51$ & $20.70 \pm 0.73$ \\
KSE & $73.47 \pm 0.50$ & $39.03 \pm 1.45$ \\
clustered\_SE & $33.55^\dagger \pm 0.87$ & $1.45 \pm 0.38$ \\
$\alpha$\_clustered\_SE & $36.55 \pm 0.87$ & $1.39^\dagger \pm 0.26$ \\
EigV & $81.88^* \pm 0.80$ & $56.36 \pm 3.26$ \\
LS & $80.87 \pm 0.63$ & $58.06^* \pm 2.06$ \\
Majority Voting & $67.93 \pm 0.48$ & $35.24 \pm 2.18$ \\
Averaging & $47.14 \pm 0.75$ & $4.69 \pm 0.51$ \\
\textbf{Ours} & $\mathbf{76.09} \pm \mathbf{0.67}$ & $\mathbf{40.70} \pm \mathbf{2.35}$ \\
\hline \hline
\end{tabular}
\end{adjustbox}
\label{tab: qwen3 coqa}
\end{table}

\paragraph{Results on the reasoning model.}
Table~\ref{tab: qwen3 coqa} reports results on the reasoning model Qwen3-4B on CoQA. As noted in the main text, longer generations often differ slightly in meaning across samples, which weakens baselines that rely on discrete semantic clusters. This effect is most pronounced for the clustered variants, where performance can drop substantially, even below random guessing. Despite being influenced by these weak scores, our method remains robust: it stays close to the strongest baselines (EigV and LS) and outperforms the remaining methods.

\begin{table}[h]
\caption{Detection power (\%) at 10\% false alarm rate across different models on HaluEval under different annotations. An asterisk ($*$) indicates the best detection power in each case, while a dagger ($\dagger$) indicates the worst. Our proposed method demonstrates robust performance across all cases.}
\centering
\begin{adjustbox}{width=\linewidth}
\begin{tabular}{llcccc}
\hline \hline Annotation & Method & Llama-2-13B & Mistral-7B & Llama-3.1-8B & DeepSeek-v2-Lite \\

\hline \hline \multirow{10}{*}{ Llama } 
& SE & $41.70 \pm 1.18$ & $35.99 \pm 0.51$ & $38.40^\dagger \pm 0.68$ & $49.60 \pm 1.00$ \\
& $\alpha$SE & $67.96 \pm 1.05$ & $62.36 \pm 0.57$ & $65.56 \pm 0.83$ & $70.02 \pm 0.79$ \\
& KSE & $74.65 \pm 1.56$ & $71.43 \pm 1.21$ & $74.87 \pm 0.73$ & $76.43 \pm 0.94$ \\
& clustered\_SE & $52.58 \pm 1.61$ & $50.64 \pm 0.61$ & $51.98 \pm 0.76$ & $61.10 \pm 0.86$ \\
& $\alpha$\_clustered\_SE & $72.10 \pm 1.22$ & $68.67 \pm 0.76$ & $68.49 \pm 0.49$ & $71.36 \pm 0.83$ \\
& EigV & $75.16 \pm 1.19$ & $70.67 \pm 0.85$ & $70.24 \pm 0.59$ & $71.91 \pm 0.49$ \\
& LS & $22.48^\dagger \pm 7.73$ & $11.06^\dagger \pm 3.97$ & $39.37 \pm 3.58$ & $0.00^\dagger \pm 0.00$ \\
& Majority Voting & $72.24 \pm 1.64$ & $68.55 \pm 0.77$ & $70.98 \pm 0.88$ & $72.76 \pm 1.00$ \\
& Averaging & $71.17 \pm 1.45$ & $65.95 \pm 0.71$ & $70.39 \pm 0.80$ & $71.89 \pm 1.16$ \\
& \textbf{Ours} & $\mathbf{75.62}^* \pm \mathbf{1.27}$ & $\mathbf{74.96}^* \pm \mathbf{1.00}$ & $\mathbf{76.23}^* \pm \mathbf{0.80}$ & $\mathbf{78.12}^* \pm \mathbf{1.04}$ \\

\hline \hline \multirow{10}{*}{ RougeL } 
& SE & $41.50^\dagger \pm 0.58$ & $35.68^\dagger \pm 0.57$ & $36.88^\dagger \pm 0.58$ & $50.99 \pm 0.38$ \\
& $\alpha$SE & $66.46 \pm 0.87$ & $60.95 \pm 0.66$ & $62.97 \pm 0.67$ & $68.84 \pm 0.86$ \\
& KSE & $71.21 \pm 1.70$ & $69.90 \pm 1.07$ & $69.35 \pm 0.75$ & $72.27 \pm 1.06$ \\
& clustered\_SE & $51.03 \pm 0.84$ & $50.06 \pm 0.45$ & $51.99 \pm 0.35$ & $60.70 \pm 0.67$ \\
& $\alpha$\_clustered\_SE & $71.77 \pm 0.76$ & $67.32 \pm 0.51$ & $68.18 \pm 0.46$ & $71.32 \pm 0.45$ \\
& EigV & $71.02 \pm 1.00$ & $64.44 \pm 0.85$ & $65.39 \pm 0.66$ & $67.49 \pm 0.62$ \\
& LS & $64.81 \pm 3.87$ & $45.74 \pm 2.35$ & $64.34 \pm 1.36$ & $41.28^\dagger \pm 5.64$ \\
& Majority Voting & $74.93 \pm 0.75$ & $70.31 \pm 0.62$ & $71.88 \pm 0.32$ & $75.41 \pm 0.49$ \\
& Averaging & $75.45 \pm 0.67$ & $69.77 \pm 0.40$ & $71.75 \pm 0.68$ & $76.71 \pm 0.33$ \\
& \textbf{Ours} & $\mathbf{75.70}^* \pm \mathbf{1.19}$ & $\mathbf{71.74}^* \pm \mathbf{0.74}$ & $\mathbf{74.93}^* \pm \mathbf{0.63}$ & $\mathbf{76.79}^* \pm \mathbf{0.96}$ \\

\hline \hline
\end{tabular}
\end{adjustbox}
\label{tab: detection power halueval annotations}
\end{table}

\begin{table}[h]
\caption{AUROC (\%) across different models and datasets under Llama annotation. $*$ and $\dagger$ indicate the best and worst AUROC in each case.}
\centering
\begin{adjustbox}{width=\linewidth}
\begin{tabular}{llcccc}
\hline \hline Dataset & Method & Llama-2-13B & Mistral-7B & Llama-3.1-8B & DeepSeek-v2-Lite \\

\hline \hline \multirow{10}{*}{ CoQA }
& SE & $65.85^\dagger \pm 0.36$ & $65.78^\dagger \pm 0.54$ & $64.07^\dagger \pm 0.35$ & $69.62^\dagger \pm 1.00$ \\
& $\alpha$SE & $85.12 \pm 0.21$ & $85.28 \pm 0.30$ & $83.15 \pm 0.30$ & $87.50 \pm 0.89$ \\
& KSE & $91.82 \pm 0.38$ & $91.94 \pm 0.34$ & $90.59 \pm 0.33$ & $92.36 \pm 1.01$ \\
& clustered\_SE & $85.74 \pm 0.72$ & $86.83 \pm 0.54$ & $85.77 \pm 0.28$ & $88.53 \pm 1.12$ \\
& $\alpha$\_clustered\_SE & $89.97 \pm 0.41$ & $90.61 \pm 0.42$ & $89.43 \pm 0.30$ & $92.22 \pm 0.75$ \\
& EigV & $93.76^* \pm 0.27$ & $94.23^* \pm 0.30$ & $93.16^* \pm 0.31$ & $94.75^* \pm 0.60$ \\
& LS & $73.53 \pm 0.92$ & $77.01 \pm 1.26$ & $73.56 \pm 0.48$ & $79.78 \pm 2.60$ \\
& Majority Voting & $91.11 \pm 0.35$ & $91.65 \pm 0.32$ & $90.30 \pm 0.23$ & $92.85 \pm 0.83$ \\
& Averaging & $88.73 \pm 0.36$ & $90.11 \pm 0.33$ & $88.24 \pm 0.18$ & $91.81 \pm 0.89$ \\
& \textbf{Ours} & $\mathbf{91.91} \pm \mathbf{0.31}$ & $\mathbf{92.41} \pm \mathbf{0.19}$ & $\mathbf{91.11} \pm \mathbf{0.19}$ & $\mathbf{93.18} \pm \mathbf{0.64}$ \\

\hline \hline \multirow{10}{*}{ TriviaQA }
& SE & $83.71 \pm 0.09$ & $83.23 \pm 0.13$ & $85.67 \pm 0.17$ & $88.08 \pm 0.13$ \\
& $\alpha$SE & $91.96 \pm 0.13$ & $91.91 \pm 0.15$ & $92.25 \pm 0.16$ & $93.72 \pm 0.11$ \\
& KSE & $93.14 \pm 0.14$ & $93.49 \pm 0.17$ & $93.37 \pm 0.18$ & $95.64 \pm 0.10$ \\
& clustered\_SE & $91.06 \pm 0.14$ & $91.93 \pm 0.18$ & $92.74 \pm 0.15$ & $94.53 \pm 0.09$ \\
& $\alpha$\_clustered\_SE & $94.59 \pm 0.08$ & $95.24 \pm 0.10$ & $95.27 \pm 0.13$ & $96.06 \pm 0.08$ \\
& EigV & $95.26^* \pm 0.08$ & $95.88^* \pm 0.08$ & $95.80^* \pm 0.09$ & $96.58^* \pm 0.06$ \\
& LS & $76.36^\dagger \pm 0.44$ & $75.91^\dagger \pm 0.82$ & $80.17^\dagger \pm 0.63$ & $82.37^\dagger \pm 0.49$ \\
& Majority Voting & $94.56 \pm 0.07$ & $94.66 \pm 0.11$ & $95.11 \pm 0.10$ & $96.29 \pm 0.08$ \\
& Averaging & $93.71 \pm 0.10$ & $93.99 \pm 0.10$ & $94.98 \pm 0.13$ & $95.88 \pm 0.06$ \\
& \textbf{Ours} & $\mathbf{94.94} \pm \mathbf{0.08}$ & $\mathbf{95.60} \pm \mathbf{0.06}$ & $\mathbf{95.53} \pm \mathbf{0.11}$ & $\mathbf{96.56} \pm \mathbf{0.10}$ \\

\hline \hline
\end{tabular}
\end{adjustbox}
\label{tab: auroc of coqa and trivia qa(10 repeated exps) under llama ann}
\end{table}

\paragraph{Results under Rouge-L and Llama annotations.} The results are reported  in Table~\ref{tab: detection power halueval annotations}, Table~\ref{tab: auroc of coqa and trivia qa(10 repeated exps) under llama ann}, together with Table~\ref{tab: auroc of coqa and trivia qa(10 repeated exps)} and Figure~\ref{fig: AUROC on HaluEval with diff annotations}. For CoQA with LLaMA annotations, we set the calibration set size to $800$ because only a limited number of non-hallucinated prompts are annotated by Llama. Under Llama annotation, our method achieves the highest detection power on HaluEval (Table~\ref{tab: detection power halueval annotations}). For AUROC on CoQA and TriviaQA under Llama annotation (Table~\ref{tab: auroc of coqa and trivia qa(10 repeated exps) under llama ann}), although lexical similarity degrades substantially and can affect aggregation, our method still ranks second and remains close to the best baseline (Spectral Eigenvalue), outperforming the remaining scores. A similar pattern holds on HaluEval: under Llama annotation (Figure~\ref{fig: AUROC on HaluEval with diff annotations}), our method is again second and comparable to the best score (Kernel Semantic Entropy). Overall, these results further support the robustness of our approach: no single baseline performs well across all annotation protocols, whereas our method is consistently best or near-best.


\begin{table}[h]
\caption{AUROC (\%) on HaluEval with ROUGE-L annotation across varying $\theta$ (Fixed $\tau = 0.3$). Under each model, columns represent $\theta = 0.1 / 0.2 / 0.3$. $*$ and $\dagger$ indicate the best and worst AUROC in each setting.}
\centering
\begin{adjustbox}{width=\linewidth}
\setlength{\tabcolsep}{2pt} 
\begin{tabular}{l ccc ccc ccc ccc}
\hline \hline 
\multirow{2}{*}{Method} & \multicolumn{3}{c}{Llama-2-13B} & \multicolumn{3}{c}{Mistral-7B} & \multicolumn{3}{c}{Llama-3.1-8B} & \multicolumn{3}{c}{DeepSeek-V2} \\
\cline{2-13}
& 0.1 & 0.2 & 0.3 & 0.1 & 0.2 & 0.3 & 0.1 & 0.2 & 0.3 & 0.1 & 0.2 & 0.3 \\
\hline \hline
SE              & NA & $64.34^\dagger$ & $62.94^\dagger$ & $64.06^\dagger$ & $60.69^\dagger$ & $58.81^\dagger$ & $65.69^\dagger$ & $62.24^\dagger$ & $60.40^\dagger$ & $73.59^\dagger$ & $70.36^\dagger$ & $67.59^\dagger$ \\
$\alpha$SE      & NA & 83.43 & 82.09 & 82.60 & 80.36 & 78.89 & 83.53 & 81.80 & 80.61 & 86.54 & 84.54 & 82.59 \\
KSE             & NA & 89.41 & $88.13^*$ & 89.79 & $88.85^*$ & $87.95^*$ & 90.05 & 89.32 & $88.79^*$ & 91.65 & $90.43^*$ & $88.60^*$ \\
clustered\_SE   & NA & 78.32 & 76.03 & 82.30 & 78.04 & 75.59 & 82.75 & 78.82 & 76.96 & 87.74 & 83.49 & 80.40 \\
$\alpha$\_clustered\_SE & NA & 85.83 & 82.85 & 89.71 & 85.53 & 82.69 & 89.39 & 85.40 & 82.47 & 91.46 & 87.48 & 84.47 \\
EigV            & NA & 86.59 & 83.23 & 90.66 & 85.83 & 81.93 & 90.82 & 85.90 & 81.69 & 91.98 & 87.16 & 83.29 \\
LS              & NA & 88.77 & 87.94 & 87.43 & 85.62 & 85.20 & 90.51 & 89.23 & 87.69 & 88.45 & 85.19 & 83.06 \\
\textbf{Ours}   & NA & $\mathbf{90.18}^*$ & $\mathbf{87.79}$ & $\mathbf{91.83}^*$ & $\mathbf{88.71}$ & $\mathbf{86.67}$ & $\mathbf{92.67}^*$ & $\mathbf{90.56}^*$ & $\mathbf{88.13}$ & $\mathbf{93.13}^*$ & $\mathbf{90.27}$ & $\mathbf{87.31}$ \\
\hline \hline
\end{tabular}
\end{adjustbox}
\begin{flushleft}
\scriptsize Note: NA indicates that the model did not produce enough consistent generations to meet the 1000-sample calibration requirement at $\theta=0.1$.
\end{flushleft}
\label{tab:sensitivity_theta_appendix}
\end{table}
\newpage
\fi

\end{document}